\documentclass{article} 
\usepackage{preprint,times}

\usepackage{amsmath,amsfonts,bm}

\def\eqref#1{equation~\ref{#1}}

\def\1{\bm{1}}

\DeclareMathAlphabet{\mathsfit}{\encodingdefault}{\sfdefault}{m}{sl}
\SetMathAlphabet{\mathsfit}{bold}{\encodingdefault}{\sfdefault}{bx}{n}

\usepackage{booktabs}
\usepackage{multirow}
\usepackage{multicol}
\usepackage{graphicx}
\usepackage{hyperref}
\usepackage{wrapfig}
\usepackage{url}
\usepackage{caption}
\usepackage[table]{xcolor}
\usepackage{amsmath, amssymb}
\usepackage{tabularx}
\usepackage{array}

\title{Learning to Be Guided: Shaping Latent \\Geometry for Classifier-Free Guidance}

\author{
\textbf{Haoming Liu}$^{1,2,*}$\quad
\textbf{Junhao Zhu}$^{2,3,*}$\quad
\textbf{Shaoyi Zheng}$^{1,2,*}$\\[4pt]
\textbf{Yuanhe Guo}$^{1,2}$\quad
\textbf{Jinnuo Liu}$^{1,2}$\quad
\textbf{Shenji Wan}$^{1,2}$\quad
\textbf{Hongyi Wen}$^{1,2}$\\[6pt]
$^{1}$New York University \quad $^{2}$New York University Shanghai \quad $^{3}$Georgia Institute of Technology 
}

\definecolor{baselinegray}{gray}{0.94}

\newcommand\blfootnote[1]{%
  \begingroup
  \renewcommand\thefootnote{}%
  \footnotetext{\hspace*{-1.8em}#1}%
  \endgroup
}

\preprint
\begin{document}

\maketitle

\begin{abstract}
\blfootnote{$^{*}$Equal contribution. Correspondence to: \texttt{\{haoming.liu, hongyi.wen\}@nyu.edu}.}
Classifier-free guidance (CFG) is a standard component of conditional diffusion models, and guided generation performance has become the \textit{de facto} measure of state-of-the-art models.
Recent work on latent diffusion models seeks to learn latent spaces that are easier to model with diffusion.
However, under CFG, generation additionally depends on how conditions are arranged in the latent space.
We study this arrangement under discrete class conditions, termed \emph{class geometry}, and investigate what class geometry allows a latent space to work well with CFG.
Under a Gaussian class model, we show that shifting variance from intra- to inter-class at fixed total variance makes classes more distinguishable under noise, and derive a noisy posterior objective whose per-sample descent direction is proportional to the oracle CFG correction.
Based on the analysis, we propose Prototype Regularization (PR), a lightweight surrogate for the intractable noisy posterior objective, which applies a cosine prototype classifier to spatially pooled latents.
With a controlled training budget (20 epochs for VAE, 80 epochs for DiT), adding PR to an LDM baseline reduces guided gFID on ImageNet from 3.31 to 2.28, surpassing a DINOv3-aligned VA-VAE (2.44) with negligible overhead.
We further demonstrate that PR generalizes to pseudo-labels derived from clustering, continuous text-embedding conditions, and video latents.
Together, our findings highlight class geometry as a critical factor in guided generation and take a first step toward CFG-compatible representation learning.
Our project page is available at: \url{https://maps-research.github.io/cfg-geometry/}.
\end{abstract}

\section{Introduction}
\label{sec:intro}

Diffusion models~\citep{ho2020denoising, song2021score, lipman2023flow} have become a predominant paradigm in generative modeling, with recent advances largely driven by conditional generation on auxiliary signals such as class labels and text~\citep{rombach2022high,esser2024scaling}.
Classifier-free guidance (CFG)~\citep{ho2021classifierfree} is a standard technique for strengthening the conditioning: during sampling, a single model makes conditional and unconditional predictions, and their displacement (i.e., the CFG correction) is amplified by a guidance scale $\omega > 1$. 
CFG improves generation quality and has become a crucial component of state-of-the-art models: headline results are generally reported under CFG, usually with a large margin over unguided sampling.~Consequently, guided performance has become the \textit{de facto} measure of a conditional diffusion model.

In parallel, latent diffusion models (LDMs)~\citep{rombach2022high} have become the prevailing approach for generating high-dimensional data: an autoencoder first compresses data into a compact latent space, and a diffusion model is then trained in this latent space.
A growing line of research designs latent spaces that are easier to model, either by aligning latents with pretrained representations~\citep{yao2025reconstruction, leng2025repa} or by adopting pretrained representations directly as the latent space~\citep{zheng2026diffusion, shi2026latent}, while another line of analysis studies which properties of a latent space make it easier for diffusion models to learn~\citep{skorokhodov2025improving, zhong2026diffusing, liu2026delvinglatentspectralbiasing}.
These efforts primarily approach latent design through the lens of \emph{diffusability}, i.e., how easily a diffusion model learns the latent distribution.
Under CFG, however, the guided score for a condition $c$ takes the form $\nabla_{z_t}\log p_t(z_t) + \omega\,\nabla_{z_t}\log p_t(c\mid z_t)$, where the second term introduces a condition-dependent correction into the sampling dynamics.
Hence, diffusability alone does not fully capture how a latent space interacts with guidance. \looseness=-1

Intuitively, the correction term $\nabla_{z_t}\log p_t(c\mid z_t)$ pushes $z_t$ toward regions where $c$ is more probable, and this push is shaped by how latents of different conditions are arranged relative to one another.
More precisely, since $p_t(c\mid z_t)$ is evaluated on noisy latents, what matters is how distinguishable this arrangement remains under noise corruption.
To characterize this arrangement more explicitly, we ground our analysis in discrete class conditions, where it reduces to the layout of classes, termed \emph{class geometry}.
We thus ask: \emph{what class geometry allows a latent space to work well with CFG?}

We address this question by analyzing the CFG correction term through the lens of class geometry (Sec.~\ref{sec:derivation}).
Specifically, under a Gaussian class model, we derive the oracle CFG correction in closed form: it contrasts the target class mean against competing class means, weighted by their noisy posteriors.
We then show that, at fixed total variance, reallocating variance from intra-class to inter-class makes classes more distinguishable under noise, so the target class posterior concentrates earlier in denoising.
Finally, we express this geometric preference as a cross-entropy objective on the noisy class posterior, whose per-sample descent direction is proportional to the oracle correction.

Guided by our analysis, we propose Prototype Regularization (PR), a tractable surrogate for the~posterior objective on clean latents (Sec.~\ref{sec:pr}).
In practice, PR applies a cosine prototype classifier to spatially pooled latents, and we show that its gradient forms a target-versus-competitor contrast analogous to that in the oracle CFG correction.
On ImageNet class-conditional generation, adding PR to an LDM baseline reduces guided gFID from 3.31 to 2.28, outperforming DINOv3-aligned~\citep{simeoni2025dinov3} VA-VAE (2.44)~\citep{yao2025reconstruction} with negligible overhead (Sec.~\ref{subsec:exp_gt}).

To isolate the source of this gain, we transplant PR class centers into LDM latents and recover most of PR's guided improvement, indicating that the gain stems largely from class geometry (Sec.~\ref{subsec:exp_transplant}).
We further demonstrate the versatility of PR through extensions to pseudo-labels, video latents, and continuous text conditions (Sec.~\ref{subsec:exp_dino}, \ref{subsec:exp_video}, \ref{subsec:exp_continuous}). Since PR relies only on discrete partitions to promote a cluster geometry, these results suggest that it applies beyond ground-truth labels whenever this geometry stays informative under noise.
Together, our work highlights class geometry as a critical factor in guided generation and takes a first step toward CFG-compatible representation learning.

Our contributions are threefold:
\begin{itemize}
  \item We analyze CFG through the lens of class geometry. Under a Gaussian class model, we derive the closed-form oracle correction, show that shifting variance from intra- to inter-class at fixed total variance improves class distinguishability under noise, and formulate a noisy posterior objective whose per-sample descent direction is proportional to this correction.
  \item We propose Prototype Regularization (PR), a lightweight surrogate for the noisy posterior objective that shapes class geometry. By classifying clean latents against prototypes, PR yields gradients analogous to the target-versus-competitor contrast in the oracle correction.
  \item We show that PR generalizes across label sources, modalities, and condition types, suggesting its effectiveness as a CFG-compatible geometry regularizer over discrete partitions. 
\end{itemize}

\section{From CFG Geometry to Representation Learning}
\label{sec:derivation}

\subsection{Preliminaries}
\label{subsec:preliminaries}

\textbf{Latent Diffusion Models.}
A latent diffusion model (LDM)~\citep{rombach2022high} operates in a compressed latent space learned by an autoencoder.
An encoder $\mathcal{E}_\theta$ maps the raw data $x\in\mathbb{R}^{D}$ to a compact latent code $z=\mathcal{E}_\theta(x)\in\mathbb{R}^{d}$ with $d\ll D$, and a decoder $\mathcal{D}_\phi$ reconstructs $\hat{x}=\mathcal{D}_\phi(z)$.
For images, the autoencoder is trained with $\mathcal{L}_{\mathrm{LDM}}
  = \mathcal{L}_{\mathrm{Recon}}
  + \lambda_{\mathrm{KL}}\,\mathcal{L}_{\mathrm{KL}}
  + \lambda_{\mathrm{LPIPS}}\,\mathcal{L}_{\mathrm{LPIPS}}
  + \lambda_{\mathrm{GAN}}\,\mathcal{L}_{\mathrm{GAN}}$.
The diffusion model is then trained in the resulting latent space, where the latents are typically standardized via scalar factors or per-channel statistics.

\textbf{Rectified Flow.} 
Rectified flow~\citep{liu2023flow,lipman2023flow} is a widely adopted training objective for diffusion models.
For a clean latent $z\in\mathbb{R}^{d}$, we construct the linear interpolant $z_t = t\,z + (1{-}t)\,\epsilon$, with independent noise $\epsilon\sim\mathcal{N}(0,I),\, t\in(0,1)$, and train a neural network $v_\psi(z_t,t)$ to predict the velocity target $v = z - \epsilon$ by minimizing $\mathcal{L}_{\mathrm{RF}}
  = \mathbb{E}_{z,\,\epsilon,\,t}
    \bigl[\lVert v_\psi(z_t,t) - (z-\epsilon) \rVert^2\bigr]$.

\textbf{Classifier-Free Guidance.}
Classifier-free guidance (CFG)~\citep{ho2021classifierfree} amplifies the effect of conditioning during sampling to obtain higher-fidelity generations.
A single network is trained with condition dropout to produce both a conditional prediction $\widehat v_c$ and an unconditional prediction $\widehat v_u$.
The guided velocity is $\widehat v_{\mathrm{cfg}}=\widehat v_u+\omega\,(\widehat v_c-\widehat v_u)$, where $\omega > 1$ is the guidance scale.

\begin{figure}[t!]
  \centering
  \includegraphics[width=0.98\linewidth]{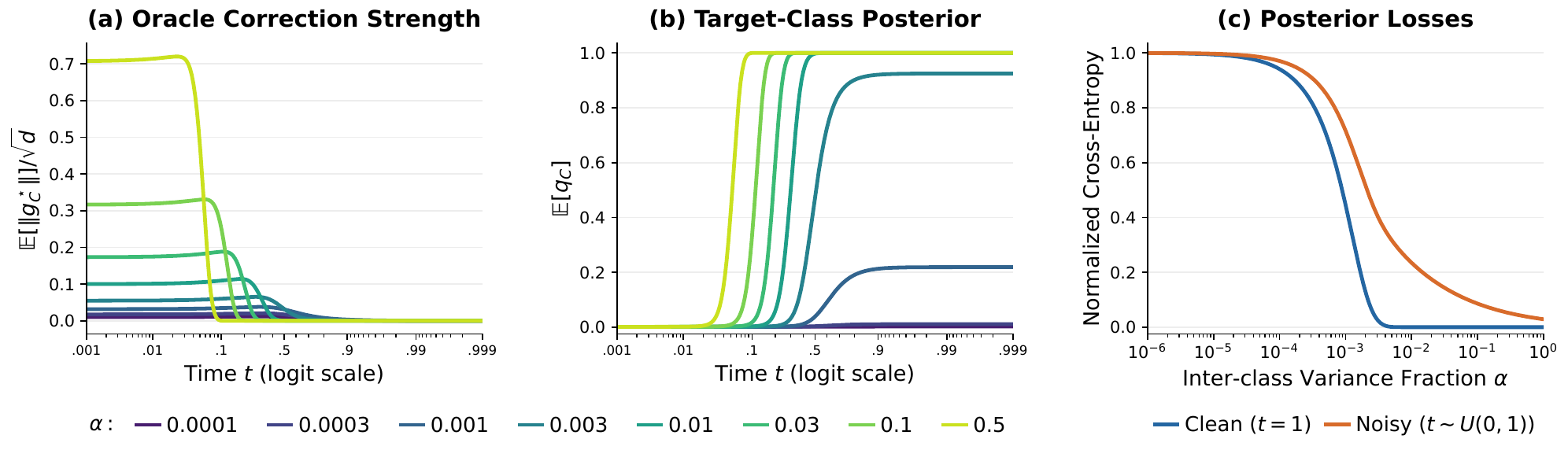}
  \vspace{-4pt}
  \caption{Simulations under the Gaussian class model with $K=1000$ classes and latent dimensions $d=8192$ (Eq.~\ref{eq:variance_allocation}).
(a) Oracle CFG correction strength along time $t$ ($t=1$ is clean), for varying inter-class variance fraction $\alpha$. The correction vanishes once the target-class posterior concentrates.
(b) Monte Carlo average of the analytic target-class posterior under forward corruption. Larger $\alpha$ yields earlier posterior concentration.
(c) Posterior losses (normalized by $\log K$) versus $\alpha$, on clean-only and uniform-time corrupted latents. Both posterior losses are non-increasing in $\alpha$.}
  \label{fig:method_3panel}
\end{figure}

\subsection{A Geometric View of Classifier-Free Guidance}
\label{subsec:cfg_geometry}

\textbf{How Class Geometry Enters CFG.}
Under rectified flow~\citep{liu2023flow}, let $v_c^\star=\mathbb{E}[v\mid z_t,t,C=c]$ and $v_u^\star=\mathbb{E}[v\mid z_t, t]$ be the Bayes-optimal conditional and unconditional predictors.
Substituting $v=(z-z_t)/(1-t)$, the oracle CFG correction $g_c^\star(z_t,t)$ reduces to:
\begin{equation}
  g_c^\star(z_t,t)
  := v_c^\star - v_u^\star
  = \frac{
      \mathbb{E}[z\mid z_t,t,C=c] - \mathbb{E}[z\mid z_t, t]
    }{1-t}.
  \label{eq:cfg_posterior_mean}
\end{equation}
Overall, the correction captures how conditioning on the target class changes the posterior estimate of the clean latent.
To gain concrete intuition on how class geometry shapes $g_c^\star$, we derive a closed form of Eq.~\ref{eq:cfg_posterior_mean} under a simplified setting.
Suppose the condition $c\in\{1,\dots,K\}$ indexes $K$ discrete classes, each with a Gaussian class-conditional prior $p(z\mid c)=\mathcal{N}(z;\,\mu_c,\,\sigma^2 I)$ with shared within-class variance $\sigma^2$, which we refer to as the \emph{Gaussian class model}.
Let $q_j=\Pr(C=j\mid z_t,t)$ denote the posterior class probability. We obtain:
\begin{equation}
g_c^\star(z_t,t)
=\frac{1-t}{(1{-}t)^2+t^2\sigma^2}
\sum_{j\neq c}q_j\,(\mu_c-\mu_j).
\label{eq:cfg_class_geometry}
\end{equation}
Therefore, the oracle CFG correction can be expressed as a posterior-weighted sum of displacements from the competing class means toward the target mean, with $\sigma^2$ shaping both the prefactor and the posterior weights.
The center displacements $\mu_c - \mu_j$ define the class contrasts, while the posterior weights $q_j$ determine which contrasts are emphasized at each noisy state $z_t$.
When the target class dominates the posterior (i.e., $q_c\to1$), the weights of all competing classes vanish, eliminating the target-versus-competitor contrast.
This behavior is illustrated in Fig.~\ref{fig:method_3panel}(a) and Fig.~\ref{fig:method_3panel}(b).

\textbf{Class Distinguishability under Noise.}
To understand how class geometry governs this posterior concentration, we examine class distinguishability under noise corruption.
Under the same Gaussian model, the class-conditional distribution of $z_t$ is $p(z_t\mid c)=\mathcal{N}\bigl(z_t;t\mu_c,(t^2\sigma^2+(1-t)^2)I\bigr)$, so the squared standardized distance between classes $c$ and $j$ is given by:
\begin{equation}
\delta_{cj}^2(t)
=
\frac{t^2\|\mu_c-\mu_j\|^2}
{t^2\sigma^2+(1-t)^2}.
\label{eq:noisy_class_separation}
\end{equation}
To isolate the effect of class geometry from changes in latent scale, we consider a controlled family of latent distributions.
With class priors held fixed, we choose the reference class means satisfying $\mathbb{E}_C[\mu_C^{\mathrm{ref}}]=0$ and $\mathbb{E}_C\|\mu_C^{\mathrm{ref}}\|^2=d$, where $d$ is the latent dimension, and set:
\begin{equation}
\mu_c(\alpha)=\sqrt{\alpha}\,\mu_c^{\mathrm{ref}},
\qquad
\sigma^2(\alpha)=1-\alpha,
\qquad
\alpha\in(0,1).
\label{eq:variance_allocation}
\end{equation}
Under this controlled family, the total variance remains $d$, with inter-class and intra-class contributions of $\alpha d$ and $(1-\alpha)d$, respectively.
Increasing $\alpha$ thus reallocates variance from intra-class to inter-class, scaling all $\delta_{cj}^2(t)$ by a common factor at every $t\in(0,1)$, so the noisy classification problem only becomes easier.
Classes therefore remain distinguishable under stronger corruption, and the target-class posterior concentrates earlier along the denoising process, as illustrated in Fig.~\ref{fig:method_3panel}(b).

\subsection{A CFG-Motivated Representation Objective}
\label{subsec:cfg_objective}

Motivated by how CFG contrasts classes through the noisy posterior probabilities in Eq.~\ref{eq:cfg_class_geometry}, we seek representations that make the target class more distinguishable from its competitors under corruption.
We express this geometric preference through an ideal posterior cross-entropy objective:
\begin{equation}
\mathcal L_{\mathrm{post}}(t)
=
\mathbb E_{z,c,\epsilon}
\left[-\log q_c\bigl(tz+(1-t)\epsilon,t\bigr)\right]
=
H(C\mid Z_t),
\label{eq:posterior_objective}
\end{equation}
where $q_c$ is the true noisy class posterior, and $H(C\mid Z_t)$ denotes the conditional entropy of the class label given the corrupted latent.
This objective is directly related to the oracle correction $g_c^\star$ via its input gradient.
With the posterior function held fixed, the per-sample descent direction satisfies:
\begin{equation}
\nabla_z\log q_c\bigl(tz+(1-t)\epsilon,t\bigr)
=
t\nabla_{z_t}\log q_c(z_t,t)
=
\frac{t^2}{1-t}\,g_c^\star(z_t,t), \quad t\in(0,1).
\label{eq:posterior_objective_gradient}
\end{equation}
Therefore, the ideal posterior objective provides a per-sample latent descent direction proportional to the oracle CFG correction.
Note that Eq.~\ref{eq:posterior_objective} is defined at a single noise level, whereas diffusion training and sampling span all noise levels. We therefore average it over a chosen time distribution $\rho$, obtaining a family of posterior objectives $\mathcal{L}_{\rho}=\mathbb{E}_{t\sim\rho}[\mathcal{L}_{\mathrm{post}}(t)]$.
Fig.~\ref{fig:method_3panel}(c) plots the losses for two choices of $\rho$ against $\alpha$: the posterior loss on clean latents ($t=1$) and its average over $t\sim\mathcal U(0,1)$.
Both losses are divided by $\log K$ (i.e., the entropy of a uniform distribution over the $K$ classes), so a value of one corresponds to complete class uncertainty in this equal-prior setting.

Within the family of class geometries in Eq.~\ref{eq:variance_allocation}, every $\mathcal{L}_\rho$ is non-increasing in $\alpha$ (Appendix~\ref{app:posterior_objective}): lower loss corresponds to larger inter-class separation and smaller intra-class spread, which keep classes distinguishable under stronger noise.
Both losses in Fig.~\ref{fig:method_3panel}(c) decrease accordingly, with the clean loss saturating at a smaller $\alpha$, while the noise-averaged loss favors further separation more persistently.
In line with this, we empirically find that uniform-time supervision can over-regularize the representation (e.g., over-smoothing its spatial variation), whereas clean-only supervision is sufficient to yield notable gains in guided generation (Appendix~\ref{app:pr_classifier_ablation}).

In summary, we obtain a family of posterior objectives whose per-sample descent directions follow the oracle CFG correction (Eq.~\ref{eq:posterior_objective_gradient}).
Our fixed-variance analysis further motivates minimizing these objectives to promote class separation and compactness.
To translate these objectives into practical regularizers, one needs to choose $\rho$ and a tractable approximation of $q_c$.

\section{Prototype Regularization}
\label{sec:pr}

Sec.~\ref{subsec:cfg_objective} motivates $\mathcal{L}_\rho$ as ideal objectives for CFG-compatible representations, but they cannot be optimized directly: the true posterior $q_c$ is defined by the very latent distribution being learned.
Since the diffusion model is trained in the latent space fixed by the autoencoder, we instead optimize a practical surrogate during autoencoder training, termed \emph{Prototype Regularization} (PR).
PR addresses both requirements identified above: it adopts clean-only supervision ($t=1$), and it approximates $q_c$ with a lightweight prototype classifier that mirrors the posterior of a Gaussian class model.

\textbf{Prototype classifier on pooled latents.}
Given a latent $z\in\mathbb{R}^{d_c\times h\times w}$, we obtain a global feature $\bar{z}=\mathrm{AvgPool}(z)\in\mathbb{R}^{d_c}$ and associate each class $j$ with a learnable prototype $p_j\in\mathbb{R}^{d_c}$, a proxy for its class mean.
With $\tilde{z}=\bar{z}/\|\bar{z}\|$ and $\tilde{p}_j=p_j/\|p_j\|$, the prototype posterior and the PR loss are:
\begin{equation}
\pi_j(z)=\frac{\exp(\tilde{z}^{\top}\tilde{p}_j/\tau)}{\sum_{k=1}^{K}\exp(\tilde{z}^{\top}\tilde{p}_k/\tau)},
\qquad
\mathcal{L}_{\mathrm{PR}}=-\log\pi_c(z),
\label{eq:pr}
\end{equation}
where $c$ is the class label and $\tau>0$ is a fixed temperature.
Average pooling restricts PR to the spatial mean of $z$, leaving the residual $z-\bar{z}$ unconstrained; this avoids pulling the full latent toward a class template, which may constrain layout and reduce diversity.
Normalization makes the objective depend only on angular alignment, which excludes the trivial minimizer of inflating feature norms.

\textbf{Connection to CFG correction.}
The posterior objective in Eq.~\ref{eq:posterior_objective} is well-defined on clean latents ($t=1$), where it equals $H(C\mid Z)$, whereas the gradient identity in Eq.~\ref{eq:posterior_objective_gradient} applies for $t\in(0,1)$.
Under the Gaussian class model with equal priors and equal-norm means, the clean posterior satisfies $q_j\propto\exp(\mu_j^{\top}z/\sigma^2)$.
Its gradient $\nabla_z\log q_c=\sigma^{-2}\sum_{j\neq c}q_j(\mu_c-\mu_j)$ is the $t\to1$ limit of the posterior gradient in Eq.~\ref{eq:posterior_objective_gradient}, providing a clean-latent counterpart to the CFG-related descent direction.
PR adopts a similar softmax form on normalized features, yielding the negative gradient:
\begin{equation}
-\nabla_{\tilde{z}}\mathcal{L}_{\mathrm{PR}}
=\frac{1}{\tau}\Bigl(\tilde{p}_c-\sum_{j=1}^{K}\pi_j\tilde{p}_j\Bigr)
=\frac{1}{\tau}\sum_{j\neq c}\pi_j\,(\tilde{p}_c-\tilde{p}_j).
\label{eq:pr_gradient}
\end{equation}
This gradient shares the class-contrast form of Eq.~\ref{eq:cfg_class_geometry}, with normalized prototypes $\tilde{p}_j$ replacing class means $\mu_j$ and classifier probabilities $\pi_j$ replacing posterior probabilities $q_j$, giving greater weight to more probable competitors.
This correspondence holds at the normalized feature $\tilde{z}$; gradients to the original latent additionally pass through normalization and pooling.
Appendix~\ref{app:additional_exp} presents ablations on PR variants, such as the prototype shape, the classifier type, and supervision on noisy latents.

\textbf{Training objective.}
We train the autoencoder with:
\begin{equation}
\mathcal{L}=\mathcal{L}_{\mathrm{LDM}}+\lambda_{\mathrm{PR}}\mathcal{L}_{\mathrm{PR}},
\label{eq:pr_objective}
\end{equation}
where the reconstructive terms in $\mathcal{L}_{\mathrm{LDM}}$ retain instance-specific information alongside the class structure from PR.
Following~\citet{esser2021taming} and~\citet{rombach2022high}, $\lambda_{\mathrm{PR}}$ is set adaptively by balancing encoder last-layer gradient norms against the reconstruction term.

\textbf{Beyond class labels.}
PR only requires a partition of the data, so ground-truth labels can be replaced by pseudo-labels from clustering pretrained features, with one prototype per cluster.
Moreover, the regularization labels need not match the generation condition: once the autoencoder is trained with PR, the diffusion model can be trained on other downstream conditions, discrete or continuous.
We show that PR remains effective in these settings (Sec.~\ref{subsec:exp_dino}, \ref{subsec:exp_continuous}).

\section{Experiments}
\label{sec:experiment}

\subsection{Experimental Setup}
\label{subsec:exp_setup}

Our primary testbed is class-conditional image generation on ImageNet~\citep{deng2009imagenet} at $256 \times 256$ resolution.
We build on the VA-VAE codebase~\citep{yao2025reconstruction} and use a convolution-based encoder/decoder with a latent shape of $32 \times 16 \times 16$, paired with LightningDiT-XL/1 as the stage-2 backbone.
We use the same number of training epochs for all methods (20 epochs for stage-1 and 80 epochs for stage-2) for fair comparison.
All ImageNet experiments in the main text are trained on $8\times$ NVIDIA H100 GPUs; a subset of appendix experiments uses 2 GPUs (with 10 epochs for stage-1 and 30 epochs for stage-2).
Full hyper-parameter details are provided in Appendix~\ref{app:exp_config}.

We use generation FID (gFID)~\citep{heusel2017gans} as our primary metric, which measures how closely generated images match the training distribution.
For all results with classifier-free guidance, we report gFID at the best guidance scale $\omega^\star$, selected via a guidance scale sweep using gFID computed on 5K samples.
When comparing against baselines, we follow the standard practice and additionally report sFID, Inception Score (IS), Precision, and Recall using the ADM evaluation toolkit~\citep{dhariwal2021diffusion}.
All metrics are computed over 50K samples unless otherwise stated.
Setups for Sec.~\ref{subsec:exp_video} and Sec.~\ref{subsec:exp_continuous} are described in their respective subsections.

\subsection{Class-Conditional Image Generation with Ground-truth Labels}
\label{subsec:exp_gt}

We compare PR against LDM~\citep{rombach2022high} and VA-VAE~\citep{yao2025reconstruction}.
We trained two VA-VAE variants, aligned with DINOv3-B and DINOv3-L~\citep{simeoni2025dinov3}, respectively. Our method integrates PR into LDM training.
On reconstruction fidelity, we report PSNR, LPIPS, and rFID.
Overall, all methods perform comparably, with VA-VAE holding a marginal edge.
PR therefore does not sacrifice reconstruction quality for class geometry.
Meanwhile, PR raises Purity@10 (i.e., the fraction of each latent's 10 nearest neighbors that share its class) from $0.020$ (LDM) to $0.279$, about $4\times$ that of the strongest VA-VAE variant ($0.071$), indicating substantially more class-consistent local neighborhoods.
PR achieves this at negligible cost, whereas VA-VAE slows stage-1 training throughput by about $25\%$ due to the additional representation alignment.

On the generation side, the gains of PR concentrate on guided settings.
Without CFG, PR improves over LDM ($13.09 \rightarrow 11.59$) but trails both VA-VAE variants.
With CFG, PR takes the lead, achieving the best gFID ($2.28$), IS ($269.85$), and Recall ($0.585$) among all methods.
Moreover, the gFID ranking with CFG is consistent with the Purity ranking, while the ranking without CFG differs.
This suggests that class geometry is particularly relevant to guided generation quality, which we investigate further in Section~\ref{subsec:exp_transplant}.
Both VA-VAE and PR exhibit lower Precision and higher Recall than LDM under CFG, suggesting broader coverage of the data distribution at a cost in sample fidelity.

\begin{figure}[t!]
  \centering
  \includegraphics[width=0.98\linewidth]{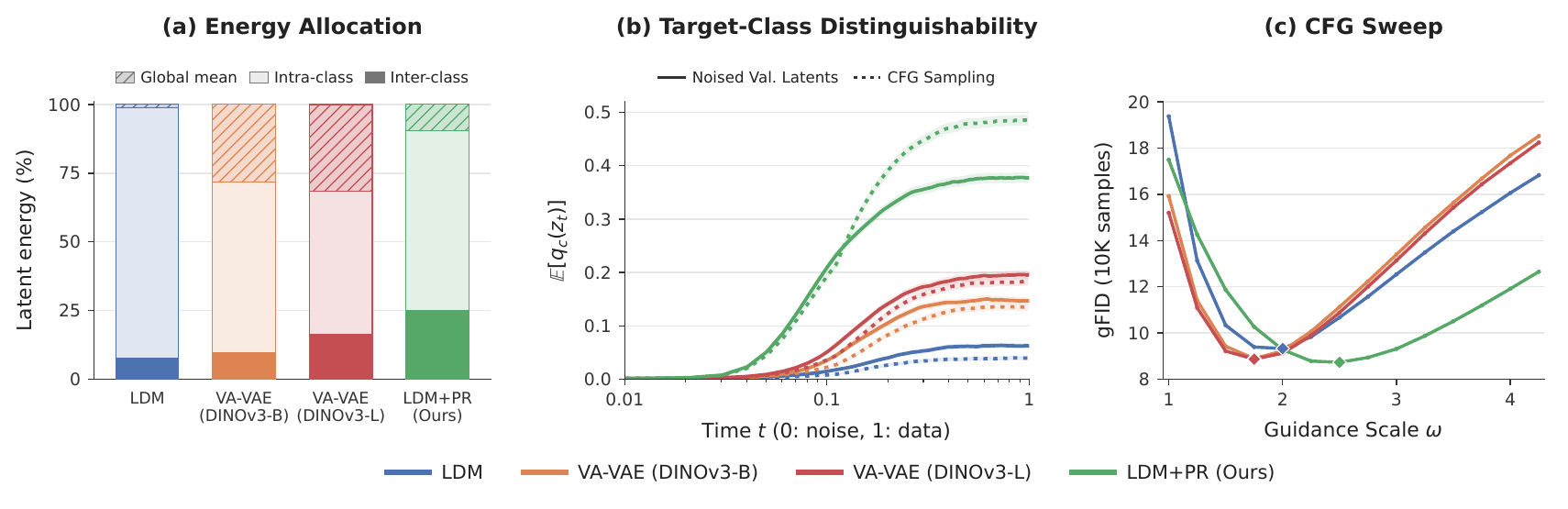}
  \vspace{-8pt}
  \caption{
(a) Energy decomposition of per-channel normalized latents into global mean $\|\mu\|^2$, inter-class variance $\mathbb{E}_c\|\mu_c-\mu\|^2$, and intra-class variance $\mathbb{E}\|z-\mu_c\|^2$.
(b) Average target-class posterior $\mathbb{E}[q_c(z_t)]$ on noised validation latents (solid) and CFG sampling trajectories (dotted), where $q_c(z_t)$ is computed analytically from the empirical distribution of ImageNet training latents under Gaussian forward corruption.
(c) gFID (10K samples) versus guidance scale $\omega$; diamonds mark the minima of the displayed curves ($\omega^\star$).
PR allocates more energy to inter-class variance, makes the target class identifiable earlier and more confidently, and reaches a lower best gFID at a larger $\omega^\star$.
}
\vspace{-8pt}
\label{fig:exp_3panel}
\end{figure}

Figure~\ref{fig:exp_3panel} examines the latent spaces from three complementary angles:
(a) We decompose the energy of per-channel normalized latents into a global mean, inter-class variation, and intra-class variation.
In LDM, over $90\%$ of the energy belongs to the intra-class portion.
VA-VAE mildly increases inter-class energy, but shifts a large portion to the global mean, a spatial template shared by all latents.
PR reallocates intra-class energy to inter-class variance, yielding the largest inter-class share among all methods.
(b) We measure the average posterior $\mathbb{E}[q_c(z_t)]$ of the target class on both forward-noised validation latents and intermediate states along CFG sampling trajectories, with $q_c(z_t)$ computed from the empirical training-latent distribution under Gaussian forward corruption.
PR yields substantially higher target-class posteriors across a broad range of noise levels, demonstrating improved class distinguishability under noise.
(c) Sweeping the guidance scale, PR reaches its optimum at a larger $\omega^\star$ and degrades more slowly beyond it, resulting in a wider near-optimal range.
This suggests that class geometry shaped via PR allows the model to benefit from stronger guidance.

\begin{table*}[t!]
\centering
\caption{
Results of class-conditional generation on ImageNet.
All quality metrics are evaluated with 50K samples.
Purity measures the fraction of each latent's $k=10$ nearest neighbors that share its class, averaged over 50K normalized validation latents.
Best results are highlighted in \textbf{bold}.
}
\vspace{-4pt}
\label{tab:main_results}

\begingroup
\small
\setlength{\tabcolsep}{5pt}
\renewcommand{\arraystretch}{1.18}

\resizebox{\textwidth}{!}{%
\begin{tabular}{l *{11}{c}}
\toprule[1.2pt]
\multirow[c]{4}{*}{\textbf{Method}} & \multicolumn{5}{c}{\textbf{Stage 1}}
& \multicolumn{6}{c}{\textbf{Stage 2}} \\
\cmidrule(lr){2-6}
\cmidrule(lr){7-12}

& \multicolumn{3}{c}{Reconstruction}
& Geometry
& Speed
& w/o CFG
& \multicolumn{5}{c}{w/ CFG} \\
\cmidrule(lr){2-4}
\cmidrule(lr){5-5}
\cmidrule(lr){6-6}
\cmidrule(lr){7-7}
\cmidrule(lr){8-12}

& rFID\,${\downarrow}$
& PSNR\,${\uparrow}$
& LPIPS\,${\downarrow}$
& Purity\,${\uparrow}$
& Steps/s\,${\uparrow}$
& gFID\,${\downarrow}$
& gFID\,${\downarrow}$
& sFID\,${\downarrow}$
& IS\,${\uparrow}$
& Precision\,${\uparrow}$
& Recall\,${\uparrow}$ \\
\midrule[0.6pt]

LDM
& 0.308 & 27.07 & 0.101 & 0.020 & \textbf{2.16}
& 13.09 & 3.31 & \textbf{4.68} & 237.73
& \textbf{0.859} & 0.504 \\

VA-VAE (DINOv3-B)
& 0.280 & \textbf{27.09} & \textbf{0.099} & 0.053 & 1.65
& 9.76 & 2.64 & 4.94 & 236.02
& 0.828 & 0.557 \\

VA-VAE (DINOv3-L)
& \textbf{0.275} & 27.06 & 0.099 & 0.071 & 1.61
& \textbf{9.01} & 2.44 & 4.83 & 242.17
& 0.816 & 0.564 \\

LDM + PR (Ours)
& 0.277 & 26.95 & 0.101 & \textbf{0.279} & 2.13
& 11.59 & \textbf{2.28} & 4.87 & \textbf{269.85}
& 0.808 & \textbf{0.585} \\

\bottomrule[1.2pt]
\end{tabular}%
}
\endgroup
\end{table*}

\begin{table}[t]
\vspace{-8pt}
\centering
\caption{Center transplantation on LDM latents ($z' = z - \mu_c^{\mathrm{LDM}} + \mu_c^{\mathrm{Ref}}$).
Rescaled LDM centers match the inter-class energy of full PR centers.
Random centers match the pooled PR energy.
Spatially constant centers are broadcast over the spatial grid.
$\omega^\star$ denotes the best CFG scale.}
\label{tab:center_transplant}
\vspace{-4pt}

\begingroup
\small
\setlength{\tabcolsep}{5pt}
\renewcommand{\arraystretch}{1.18}
\definecolor{baselinegray}{gray}{0.94}

\begin{tabular}{cccccc}
\toprule[1.2pt]
\multirow{2}{*}{\textbf{Base Latent}}
& \multirow{2}{*}{\textbf{Center Construction}}
& \multirow{2}{*}{\textbf{Center Shape}}
& \multicolumn{2}{c}{\textbf{gFID@50K}\,${\downarrow}$}
& \multirow{2}{*}{$\omega^\star$} \\
\cmidrule(lr){4-5}
& & & w/o CFG & w/ CFG & \\
\midrule[0.6pt]

\rowcolor{baselinegray}
LDM + PR
& Native & --
& 11.59 & 2.28 & 2.5 \\

\rowcolor{baselinegray}
LDM
& Native & --
& 13.09 & 3.31 & 2.0 \\

\midrule

LDM
& LDM (Rescaled) & $32\times16\times16$
& 12.58 & 2.89 & 2.0 \\

LDM
& PR & $32\times16\times16$
& 12.79 & 2.63 & 3.5 \\

LDM
& PR & $32\times1\times1$
& 12.65 & 2.52 & 4.0 \\

LDM
& Random Directions + PR Norms & $32\times1\times1$
& 13.74 & 3.16 & 4.5 \\

\bottomrule[1.2pt]
\end{tabular}
\endgroup
\vspace{-6pt}
\end{table}

\subsection{Understanding the Gains through Center Transplantation}
\label{subsec:exp_transplant}

Figure~\ref{fig:exp_3panel}(a) shows that PR primarily reallocates latent energy to inter-class variance (i.e., more separated class centers).
This raises a natural question: \emph{can an unregularized latent space benefit from the class geometry of PR in a plug-and-play manner?}
To answer this, we transplant reference class centers into the LDM latent space.
For a latent $z$ of class $c$, we replace its native center with a reference center, $z' = z - \mu_c^{\mathrm{LDM}} + \mu_c^{\mathrm{Ref}}$, which preserves intra-class deviations, and train a stage-2 model on $z'$.
To ensure the decoder receives latents in its native space, we apply the inverse mapping before decoding.
We use two PR references: the full center and its spatially pooled version.
As controls, we use LDM's own centers rescaled to match the inter-class energy of the full PR centers, and random directions with PR's per-class norms.
As presented in Tab.~\ref{tab:center_transplant}, transplanting PR centers improves guided generation (gFID $3.31 \rightarrow 2.52$ with pooled centers), while the unguided setting changes only marginally ($13.09 \rightarrow 12.65$).
Increasing the inter-class energy of LDM alone yields a partial gain ($2.89$), and PR centers improve further beyond this energy-matched control.
In contrast, random directions provide little benefit with CFG and even hurt unguided generation.
These results indicate that PR centers have a better arrangement, and the gain does not stem from increased inter-class energy alone.
Besides, pooled centers outperform spatially resolved ones ($2.52$ vs.\ $2.63$), so the useful class signal is mainly a global channel-wise offset.

Interestingly, transplanting PR or random centers shifts $\omega^\star$ upward (3.5 to 4.5), whereas rescaling LDM’s own centers does not (2.0). However, a larger $\omega^\star$ does not by itself imply better geometry: random centers admit the strongest guidance yet bring little gain in guided quality, while rescaled LDM centers improve quality without changing $\omega^\star$.
Hence, the optimal guidance scale and the quality attained at it should be considered separately.
Finally, center transplantation recovers most of PR's guided gain but only a small portion of its unguided gain, and thus does not fully reproduce the benefits of PR.
This is likely because transplantation only alters inter-class separation and leaves intra-class spread untouched, whereas PR reshapes both during stage-1 training.

\subsection{Generalization under Pseudo-Labels and Non-identical Label Sets}
\label{subsec:exp_dino}
\vspace{-4pt}

\noindent
\begin{minipage}[t]{0.5\textwidth}
\vspace{0pt}
Building on PR's performance gains on ground-truth class labels, we further investigate whether PR remains effective on other types of labels, and whether the two stages must share the same label set.
To this end, we construct unsupervised labels by clustering the \texttt{[CLS]} tokens of DINOv3-L~\citep{simeoni2025dinov3} with spherical $K$-means, setting $K{=}1000$ to match the number of ImageNet classes.
We then evaluate all combinations of stage-1 PR labels and stage-2 conditioning labels (Tab.~\ref{tab:dino_labels}).
When conditioning on DINO clusters, we sample cluster labels in proportion to cluster sizes during evaluation.
\end{minipage}%
\hfill
\begin{minipage}[t]{0.48\textwidth}
\centering
\captionof{table}{PR with different label set pairings for stage-1 regularization and stage-2 conditioning.
We report guided gFID $/$ unguided gFID below. The best result per-column is in \textbf{bold}.}
\label{tab:dino_labels}
\vspace{-2pt}
\begingroup
\small
\setlength{\tabcolsep}{5pt}
\renewcommand{\arraystretch}{1.18}
\resizebox{\linewidth}{!}{%
\begin{tabular}{lcc}
\toprule[1.2pt]
& \multicolumn{2}{c}{\textbf{Stage 2 Condition}} \\
\cmidrule(lr){2-3}
\textbf{PR Labels} & Ground-truth & DINO Clusters \\
\midrule[0.6pt]
\rowcolor{baselinegray}
None & 3.31 / 13.09 & 3.58 / 13.31 \\
\midrule
Ground-truth & \textbf{2.28} / 11.59 & \textbf{2.42} / 11.82 \\
DINO Clusters & 2.34 / \textbf{11.44} & 2.55 / \textbf{11.47} \\
\bottomrule[1.2pt]
\end{tabular}%
}
\endgroup
\end{minipage}

Using ground-truth labels in both stages yields the best guided generation ($2.28$).
Replacing the stage-1 labels with DINO clusters recovers most of this gain ($2.34$, about $94\%$ of the improvement over the no-PR baseline) and even improves unguided generation ($11.44$ vs.\ $11.59$).
When stage-2 conditions on DINO clusters, PR also substantially improves over the no-PR baseline ($3.58 \rightarrow 2.42$ and $2.55$ with CFG).
Notably, under DINO conditioning, PR regularized with ground-truth labels performs best with CFG ($2.42$), even though its stage-1 labels differ from the stage-2 condition.
These results show that PR neither relies on ground-truth labels nor requires the two stages to share the same label set, making it applicable to settings without human annotations.

\subsection{Generalization to Video Latents}
\label{subsec:exp_video}

We further examine whether PR generalizes to video latents.
We conduct experiments on UCF-101~\citep{soomro2012ucf101}, a human action video dataset spanning 101 action classes, and adopt LeanVAE-16~\citep{cheng2025leanvae} as the base autoencoder.
This VAE is trained with reconstruction objectives and compresses each 17-frame $256\times256$ clip into a latent of shape $5\times16\times32\times32$ ($T\times C\times H\times W$).
Starting from the official VAE checkpoint, we fine-tune for 3 epochs using either reconstruction alone or reconstruction with PR; for video latents, PR's average pooling spans both the spatial and temporal dimensions.
We then fine-tune a Latte-XL/2 model~\citep{ma2025latte} on each of three latent spaces: the official VAE, the VAE fine-tuned with reconstruction alone, and the VAE fine-tuned with PR.
All three Stage 2 runs are initialized from the EMA weights of the same Latte-XL/2 checkpoint provided by LeanVAE and fine-tuned for 50K steps with a global batch size of 10.
We also evaluate the VAE and Latte-XL/2 checkpoints provided by LeanVAE without further training as a reference.
We report FVD (over 2048 clips), which measures the distance between the generated and real video distributions, and action classification accuracy from a pre-trained UCF-101 classifier, which measures the class consistency of conditional generation.
As shown in Tab.~\ref{tab:video_generalization}, PR improves FVD in both settings with on-par accuracy, reducing FVD from $152.7$ to $140.1$ with CFG and from $287.3$ to $275.7$ without CFG.
These results suggest that the benefit of PR is not specific to images and extends to latent spaces of other modalities.

\begin{table}[t]
\centering
\caption{Generalization to video latents on UCF-101~\citep{soomro2012ucf101} using LeanVAE-16~\citep{cheng2025leanvae} and Latte-XL/2~\citep{ma2025latte}.
\emph{Official} denotes the VAE and Latte-XL/2 checkpoints provided by LeanVAE, used without further training.
We report FVD and action classification accuracy with and without CFG.
The best results are in \textbf{bold}.}
\label{tab:video_generalization}
\vspace{-4pt}

\begingroup
\small
\setlength{\tabcolsep}{5pt}
\renewcommand{\arraystretch}{1.18}

\begin{tabular}{cccccc}
\toprule[1.2pt]
\multirow{2}{*}{\shortstack{\textbf{Stage 1}\\\textbf{(LeanVAE)}}}
& \multirow{2}{*}{\shortstack{\textbf{Stage 2}\\\textbf{(Latte-XL/2)}}}
& \multicolumn{2}{c}{\textbf{w/ CFG}}
& \multicolumn{2}{c}{\textbf{w/o CFG}} \\
\cmidrule(lr){3-4}
\cmidrule(lr){5-6}
& &
FVD\,${\downarrow}$
& Accuracy (\%)\,${\uparrow}$
& FVD\,${\downarrow}$
& Accuracy (\%)\,${\uparrow}$ \\
\midrule[0.6pt]

\rowcolor{baselinegray}
Official
& Official
& 167.0 & 92.9 & 340.2 & 87.2 \\

\rowcolor{baselinegray}
Official
& Fine-Tuned
& 161.4 & 98.5 & 302.0 & 92.7 \\

\midrule
Fine-Tuned (Recon Only)
& Fine-Tuned
& 152.7 & \textbf{99.0} & 287.3 & \textbf{93.0} \\

Fine-Tuned (PR, Ours)
& Fine-Tuned
& \textbf{140.1} & 98.9 & \textbf{275.7} & \textbf{93.0} \\

\bottomrule[1.2pt]
\end{tabular}
\endgroup
\vspace{-2pt}
\end{table}

\begin{table*}[t]
\centering
\caption{ImageNet text-to-image generation using different Stage-1 latent representations.
All metrics use 50K images generated from held-out validation captions.
All alignment to CLIP-L adopts its text embeddings.
CLIP denotes mean cosine similarity of CLIP embeddings calculated from text captions and generated images.
All cluster labels use $K=1000$.
The best results are in \textbf{bold}.}
\label{tab:t2i_results}
\vspace{-4pt}

\begingroup
\small
\setlength{\tabcolsep}{5pt}
\renewcommand{\arraystretch}{1.18}
\definecolor{baselinegray}{gray}{0.94}

\resizebox{\textwidth}{!}{%
\begin{tabular}{lcccccc}
\toprule[1.2pt]
\multirow{2}{*}{\textbf{Method}}
& \multicolumn{3}{c}{\textbf{w/ CFG}}
& \multicolumn{3}{c}{\textbf{w/o CFG}} \\
\cmidrule(lr){2-4}
\cmidrule(lr){5-7}
& gFID Train\,${\downarrow}$
& gFID Val\,${\downarrow}$
& CLIP\,${\uparrow}$
& gFID Train\,${\downarrow}$
& gFID Val\,${\downarrow}$
& CLIP\,${\uparrow}$ \\
\midrule[0.6pt]

\rowcolor{baselinegray}
LDM
& 3.15 & 2.90 & 0.2775
& 10.38 & 7.52 & 0.2577 \\

\midrule
VA-VAE (DINOv3-L)
& 2.53 & 2.67 & 0.2814
& \textbf{8.02} & \textbf{5.65} & \textbf{0.2620} \\

Cosine Alignment (CLIP-L)
& 2.93 & 3.04 & 0.2799
& 9.52 & 6.84 & 0.2590 \\

Cosine Alignment (DINOv3-L)
& 2.86 & 2.78 & 0.2784
& 9.35 & 6.74 & 0.2585 \\

\midrule
LDM + PR (GT Labels)
& \textbf{2.51} & \textbf{2.35} & 0.2811
& 9.63 & 6.87 & 0.2570 \\

LDM + PR (CLIP-L Clusters)
& 2.52 & 2.42 & 0.2813
& 10.15 & 7.36 & 0.2578 \\

LDM + PR (DINOv3-L Clusters)
& 2.53 & 2.46 & \textbf{0.2818}
& 9.41 & 6.72 & 0.2575 \\

\bottomrule[1.2pt]
\end{tabular}
}
\endgroup
\vspace{-8pt}
\end{table*}

\subsection{Generalization to Continuous Conditioning}
\label{subsec:exp_continuous}

Finally, we examine whether PR benefits generation under continuous conditioning signals.
We consider text-to-image generation on ImageNet: we caption each image with Florence-2-large~\citep{xiao2024florence} and encode the caption into a CLIP ViT-L/14 text embedding~\citep{radford2021clip}, which replaces the class embedding in LightningDiT~\citep{yao2025reconstruction} through a small MLP adapter.
Besides LDM and VA-VAE (DINOv3-L), we compare against two continuous alignment baselines, which align latents via cosine similarity to pre-trained features projected to the channel dimension by PCA.
We apply 1000-class PR with ground-truth classes, clusters of CLIP-L text embeddings, and DINOv3-L \texttt{[CLS]} tokens.
All metrics are computed on 50K images generated from held-out validation captions.
As shown in Tab.~\ref{tab:t2i_results}, PR variants lead on all three metrics under guidance, with cluster labels consistently outperforming cosine alignment to the same representation.
Meanwhile, clusters of CLIP-L text or DINOv3-L image embeddings yield performance comparable to ground-truth classes.
Without CFG, VA-VAE performs best, consistent with its leading gFID in Tab.~\ref{tab:main_results}.
Since captions typically group around shared objects or scenes, the latent geometry induced by PR may help organize representations along semantic groupings reflected in the captions, making it informative for text-conditioned generation.
Together, these results demonstrate that PR improves guided text-to-image generation without requiring ground-truth class labels.

\section{Related Work}
\label{sec:related_work}

\textbf{Latent Representation for Diffusion Models.}
Latent diffusion models (LDMs) adopt a two-stage paradigm~\citep{rombach2022high}: an autoencoder maps high-dimensional data to a compact latent space, and a diffusion model learns to generate within it.
Subsequent work expands the design space by widening the latent channels \citep{esser2024scaling,labs2025flux} and raising the compression ratio \citep{chen2025deep}.
Recent work focuses on improving latent representations.
One line of work uses pretrained representations as supervision: VA-VAE \citep{yao2025reconstruction} aligns latents with pretrained features during autoencoder training; REPA-E \citep{leng2025repa} enables end-to-end tuning through REPA alignment \citep{yu2024representation}; MAETok \citep{chen2025masked} uses masked modeling with auxiliary decoders.
Another line builds on pretrained encoders: AlignTok \citep{chen2025aligntok} adapts a pretrained encoder through multi-stage training; RAE \citep{zheng2026diffusion,singh2026improved} uses frozen pretrained representations as diffusion latents; and SVG \citep{shi2026latent} augments frozen DINO features with learned residual features.
VTP \citep{yao2026towards} further explores joint training objectives, combining image-text contrastive, self-supervised, and reconstruction losses.

\textbf{Diffusability of Latent Representations.}
Recent work studies latent \emph{diffusability} (i.e.,  how easily diffusion models learn a latent distribution).
Some approaches focus on spectral structure and spatial equivariance: \citet{skorokhodov2025improving} use scale-equivariance regularization to suppress excessive high-frequency content; EQ-VAE \citep{kouzelis2025eqvae} promotes equivariance by matching reconstructions of transformed latents to transformed images; SSVAE \mbox{\citep{liu2026delvinglatentspectralbiasing}} studies latent spectral properties in video.
Others regularize the latent distribution directly: UL \citep{heek2026unified} uses a learned diffusion prior; DMVAE \citep{ye2025distribution} aligns the aggregate latent distribution with a reference via score-based distribution matching.
Complementary studies examine multiple properties jointly: PAE \citep{yue2026matters} highlights spatial structure, local manifold continuity, and global semantics; \citet{zhong2026diffusing} identify velocity irreducible variance, semantic separability, and spatial structure as consistent predictors of generation quality in a large-scale study.

\textbf{Classifier-Free Guidance.}
Classifier-free guidance (CFG) \citep{ho2021classifierfree} strengthens conditioning at inference at the cost of reduced diversity, with subsequent works refining this trade-off in sampling \citep{karras2024guiding,kynkaanniemi2024applying}.
Analyses show that CFG does not sample from the intended tilted distribution \citep{bradley2024classifier,chidambaram2024does} and raises classification confidence while reducing diversity \citep{wu2024theoretical}. 
It can be decomposed into a class-mean shift and contrastive components in linear models \citep{li2025towards}, or viewed as pushing trajectories away from decision boundaries \citep{zhao2026studying}.
\citet{pavasovic2025classifier} show that in high dimensions CFG acts mainly during class selection, consistent with the vanishing correction in Fig.~\ref{fig:method_3panel}(a).
These works primarily analyze CFG under a given data distribution; we instead treat the latent space as a design variable and study which class geometry facilitates CFG.

\textbf{Class Separation for Guided Generation.}
Recent work links CFG to class separation: \citet{ventura2026emergence} show that CFG distorts the conditional target when classes are poorly separated.
MCLR \citep{li2026mclr} attributes the reliance on CFG to weak separation between learned conditionals and enlarges it by maximizing inter-class likelihood ratios when training the generator; more broadly, guidance-free training internalizes the effect of CFG into the generator \citep{chen2025visual,tang2025diffusion,chen2025toward}.
ProtoDiffusion \citep{baykal2024protodiffusion} instead modifies the conditioning, replacing learned class embeddings with pretrained class prototypes.
These methods change how the generator is trained or conditioned while keeping the data representation fixed, whereas PR reshapes the latent space itself to promote class separation in the diffusion target.

\newpage
\section{Discussion}
\label{sec:discussion}

\textbf{Unguided performance.}
PR primarily targets guided generation, where its gains concentrate; without guidance, it still improves over LDM but trails VA-VAE (Tab.~\ref{tab:main_results}).
The remaining gap may stem from average pooling, which constrains only the channel dimension and leaves spatial structure unregularized, while suppressing high-frequency spatial components has been shown to be helpful for diffusability~\citep{skorokhodov2025improving, liu2026delvinglatentspectralbiasing}.
Since PR leaves the spatial residual unconstrained, it can be combined with regularizers that shape other aspects of the latent space: applying PR on top of VA-VAE improves both guided and unguided generation (Appendix~\ref{app:compositionality}).

\textbf{Scope of analysis.}
Under Gaussian forward corruption, Tweedie's formula links the posterior gradient to the Bayes-optimal CFG correction (Eq.~\ref{eq:posterior_objective_gradient}) without assuming Gaussian latents (Appendix~\ref{app:posterior_objective_gradient}).
The Gaussian class model additionally provides a closed-form correction and predicts how variance reallocation affects noisy class separation and posterior uncertainty.
The empirical posterior trend is consistent with these predictions (Fig.~\ref{fig:exp_3panel}(b)).
Appendix~\ref{app:cfg_perspectives} further discusses how class geometry relates to other CFG quantities, complementing the posterior-based analysis.

\textbf{Design space of PR.}
PR instantiates the posterior objective through two design choices, the time distribution $\rho$ and the approximation of $q_c$, both of which leave room for exploration (Appendix~\ref{app:pr_classifier_ablation}).
For $\rho$, clean-latent supervision avoids the over-regularization of noisy supervision, though intermediate distributions may offer better trade-offs.
For $q_c$, the pooled prototype classifier is one of many ways to realize the target-versus-competitor contrast.

\section{Conclusion}
\label{sec:conclusion}

In this work, we study CFG-compatible representation learning through the lens of class geometry.
Our analysis links the oracle CFG correction to how distinguishable classes remain under noise, which motivates Prototype Regularization (PR), a lightweight regularizer that shapes class geometry and improves guided generation.
PR generalizes across label choices, data modalities, and condition signals, suggesting that aligning latent geometry through categorical priors is a simple yet effective way to design latent spaces for guided generation.
Given that CFG is used almost universally at sampling time, we hope this work encourages treating CFG compatibility as an explicit and principled axis of latent design, alongside reconstruction and generation quality.

\newpage
\section*{Acknowledgments}
This work is supported by the NYU Shanghai Center for Data Science. This work is also supported in part through the NYU IT High Performance Computing resources, services, and staff expertise.
\bibliography{preprint}
\bibliographystyle{preprint}

\newpage
\appendix

\section*{Appendix}
The appendix is organized as follows:
\begin{itemize}
    \item \textbf{Appendix~\ref{app:latent_vis}} presents UMAP visualizations of the learned VAE latents.
    \item \textbf{Appendix~\ref{app:cfg_vis}} provides additional visualizations of noisy posterior and oracle velocity field.
    \item \textbf{Appendix~\ref{app:exp_config}} provides full details of the experimental configurations.
    \item \textbf{Appendix~\ref{app:additional_exp}} reports additional experimental results.
    \item \textbf{Appendix~\ref{app:cfg_derivations}} derives the equations and claims stated in the main text.
    \item \textbf{Appendix~\ref{app:cfg_perspectives}} derives complementary perspectives on classifier-free guidance (CFG).
\end{itemize}

\newpage
\section{Latent Visualizations}
\label{app:latent_vis}

\begin{figure}[h!]
    \centering
    \includegraphics[width=0.75\textwidth]{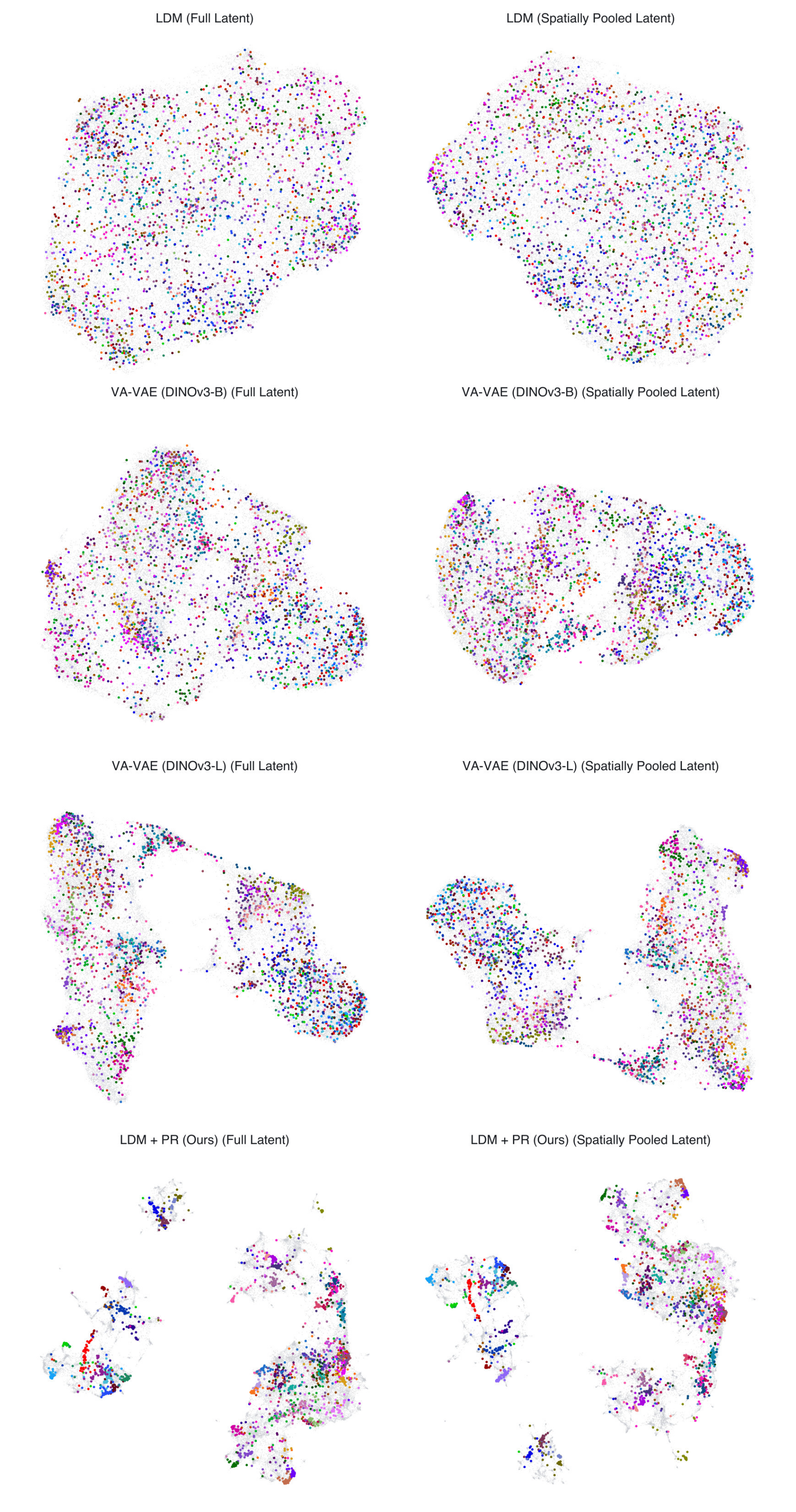}
    \caption{UMAP visualization of full and spatially pooled latents from LDM, VA-VAE, and PR (ours). We encode 50K validation images and plot latents from 50 selected classes, colored by class label. Latents are normalized per dimension before UMAP. PR produces noticeably tighter and better separated class clusters. Zoom in for the best view. Better view with colors.}
    \label{appfig:latent_vis}
\end{figure}

\newpage
\section{Classifier-Free Guidance Visualizations}
\label{app:cfg_vis}

\begin{figure}[h!]
    \centering
    \includegraphics[width=0.98\textwidth]{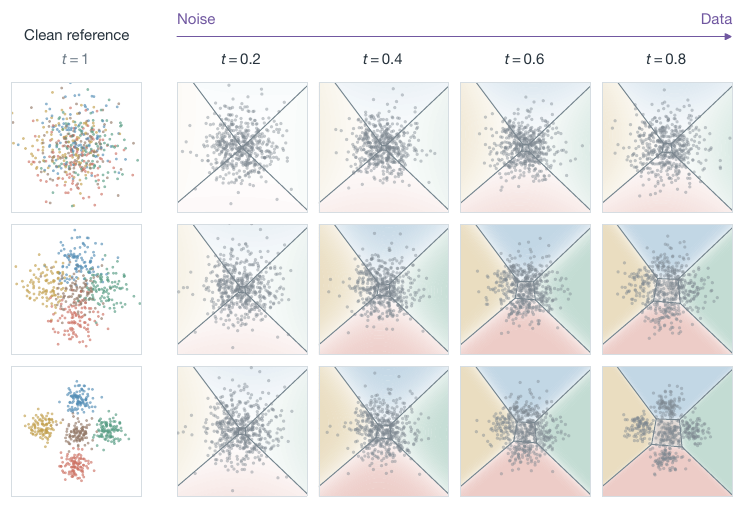}
    \caption{Noisy marginals of $z_t$ at several $t\in(0,1)$ for toy Gaussian mixtures with fixed total variance but different allocations between inter-class and intra-class variance. Background colors mark the Maximum A Posteriori (MAP) decision regions $\arg\max_c q_c(z_t)$, with opacity proportional to the confidence $\max_c q_c(z_t)$. With larger inter-class variance, classes become separable at smaller $t$, i.e., at noisier levels, and the confident MAP regions align more closely with the true class supports.}
    \label{appfig:vis_posterior}
\end{figure}

\begin{figure}[h!]
    \centering
    \includegraphics[width=0.98\textwidth]{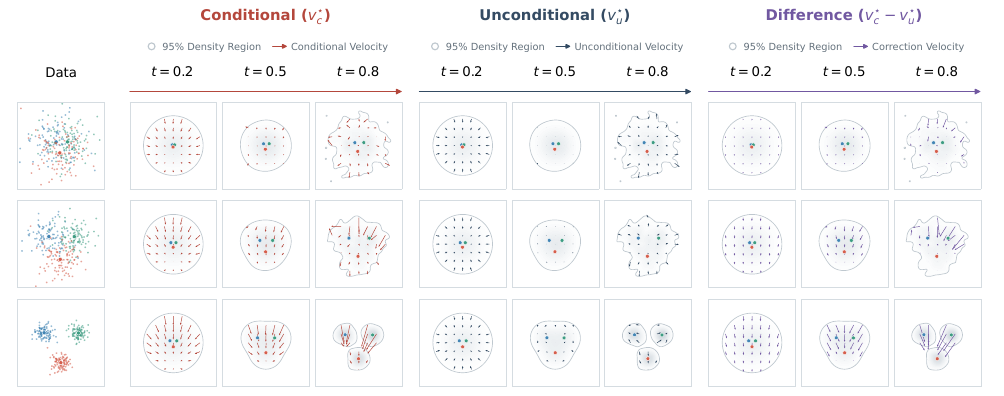}
    \caption{Oracle velocity fields on toy Gaussian mixtures with fixed total variance. From top to bottom, the fraction of variance allocated to class means increases. For each space, we show the conditional velocity $v_c^\star$ (conditioned on the red class), the unconditional velocity $v_u^\star$, and the correction $g_c^\star=v_c^\star-v_u^\star$ at $t\in\{0.2,0.5,0.8\}$. Gray contours mark the 95\% density region of $z_t$, and dots mark the class means of $z_t$. With clearer class geometry, the conditional and correction velocities point more consistently toward the target class mean, and the correction vanishes where the posterior concentrates on the target class. Zoom in for the best view. Better view with colors.}
    \label{appfig:vis_velocity}
\end{figure}

\newpage
\section{Experiment Configurations}
\label{app:exp_config}

\begin{table}[h]
\centering
\caption{Training and evaluation configurations for ImageNet experiments.
Full-budget experiments use 8 GPUs; reduced-budget ablations use 2 GPUs.
Both budgets use the same architectures and evaluation protocol,
with experiment-specific CFG search ranges listed below.}
\label{apptab:training_evaluation_configs}

\begingroup
\scriptsize
\setlength{\tabcolsep}{5pt}
\renewcommand{\arraystretch}{1.12}
\definecolor{sectiongray}{gray}{0.94}

\begin{tabularx}{\linewidth}{
    @{}l
    >{\centering\arraybackslash}X
    >{\centering\arraybackslash}X@{}
}
\toprule[1.2pt]
\textbf{Configuration}
& \textbf{Full Budget}
& \textbf{Reduced Budget} \\
\midrule[0.6pt]

\rowcolor{sectiongray}
\multicolumn{3}{l}{\textbf{Data and Architecture}} \\
Dataset
& \multicolumn{2}{c}{ImageNet, $256\times256$} \\
GPUs per Run
& 8 & 2 \\
Stage 1 Architecture
& \multicolumn{2}{c}{KL-VAE, downsampling factor 16} \\
VAE Base Channels / Multipliers
& \multicolumn{2}{c}{128 / $(1,1,2,2,4)$} \\
VAE Residual Blocks / Attention Resolution
& \multicolumn{2}{c}{2 per level / $16\times16$} \\
Latent Shape ($C\times H\times W$)
& \multicolumn{2}{c}{$32\times16\times16$} \\
Stage 2 Architecture
& \multicolumn{2}{c}{LightningDiT-XL/1} \\
DiT Depth / Hidden Dimension / Heads
& \multicolumn{2}{c}{28 / 1152 / 16} \\
DiT Patch Size
& \multicolumn{2}{c}{$1\times1$} \\
DiT Components
& \multicolumn{2}{c}{QK-Norm, SwiGLU, RoPE, RMSNorm} \\

\midrule
\rowcolor{sectiongray}
\multicolumn{3}{l}{\textbf{Stage 1 Training}} \\
Image Preprocessing
& \multicolumn{2}{c}{Shortest-side resize to 256, random crop, no flip} \\
Base Objective
& \multicolumn{2}{c}{L1 + LPIPS + KL + adaptive adversarial loss} \\
KL Weight / Adversarial Weight Factor
& \multicolumn{2}{c}{$10^{-6}$ / $0.5$} \\
Discriminator Start
& \multicolumn{2}{c}{2,500 steps} \\
Optimizer (Generator and Discriminator)
& \multicolumn{2}{c}{Adam, $(\beta_1,\beta_2)=(0.5,0.9)$} \\
Learning Rate / Schedule
& \multicolumn{2}{c}{$10^{-4}$ / constant, no warmup} \\
Batch Size per GPU
& 32 & 32 \\
Global Batch Size
& 256 & 64 \\
Training Epochs
& 20 & 10 \\
Recorded Training Steps
& 100,080 & 200,180 \\

\midrule
\rowcolor{sectiongray}
\multicolumn{3}{l}{\textbf{Stage 2 Training}} \\
Latent Extraction
& \multicolumn{2}{c}{Frozen EMA encoder, posterior mode, center crop} \\
Latent Normalization
& \multicolumn{2}{c}{Per-channel mean and standard deviation} \\
Objective / Prediction
& \multicolumn{2}{c}{Linear-path flow matching / velocity} \\
Training Time Sampler
& \multicolumn{2}{c}{Logit-normal} \\
Optimizer
& \multicolumn{2}{c}{AdamW} \\
Optimizer $(\beta_1,\beta_2)$
& $(0.9,0.95)$ & $(0.9,0.995)$ \\
Learning Rate / Schedule
& \multicolumn{2}{c}{$10^{-4}$ / constant, no warmup} \\
Weight Decay / Max Gradient Norm
& \multicolumn{2}{c}{0 / 1.0} \\
Batch Size per GPU
& 128 & 128 \\
Global Batch Size
& 1,024 & 256 \\
Training Duration
& 80 epochs & $\sim$30 epochs \\
Recorded Training Steps
& 100,080 & 150,000 \\
Condition Dropout
& \multicolumn{2}{c}{0.1} \\

\midrule
\rowcolor{sectiongray}
\multicolumn{3}{l}{\textbf{Shared Training Settings}} \\
Precision
& \multicolumn{2}{c}{BF16 mixed precision} \\
EMA Decay
& \multicolumn{2}{c}{0.9999} \\
Training Seed
& \multicolumn{2}{c}{42} \\

\midrule
\rowcolor{sectiongray}
\multicolumn{3}{l}{\textbf{Sampling and Evaluation}} \\
Evaluation Weights
& \multicolumn{2}{c}{EMA} \\
Stage 1 Evaluation
& \multicolumn{2}{c}{50K validation images; rFID, PSNR, LPIPS, L1} \\
ODE Solver / Steps
& \multicolumn{2}{c}{Euler / 50} \\
Sampling Time Grid
& \multicolumn{2}{c}{Uniform, no timestep shift} \\
CFG Interval
& \multicolumn{2}{c}{Enabled for $t\geq0.11$ ($t=0$ denotes noise)} \\
CFG Selection
& \multicolumn{2}{c}{Minimum torch-FID over 5K generated samples} \\
Default CFG Grid
& \multicolumn{2}{c}{$\{1.25,1.50,\ldots,3.50\}$ (extended by step of 0.5 if still improving)} \\
Final Generation Evaluation
& \multicolumn{2}{c}{50K samples with selected CFG; 50K without CFG} \\
Generation Metrics / Evaluator
& \multicolumn{2}{c}{FID, sFID, IS, Precision, Recall / OpenAI ADM} \\
Default FID Reference
& \multicolumn{2}{c}{ImageNet training reference batch} \\
Additional T2I Evaluation
& \multicolumn{2}{c}{Validation-reference FID and image--prompt CLIP cosine} \\

\bottomrule[1.2pt]
\end{tabularx}
\endgroup
\end{table}

\newpage
\section{Additional Experiment Results}
\label{app:additional_exp}

\subsection{Ablation on Classifier Type in PR}
\label{app:pr_classifier_ablation}

\begin{table}[h!]
\centering
\caption{\textbf{Ablation on the classifier type in PR (reduced compute budget).} We vary the classifier head (cosine vs.\ linear) and the latent on which PR is applied (clean latents only vs.\ noisy latents at uniformly sampled timesteps). The cosine head on clean-only latents achieves the best rFID and gFID, both with and without CFG. Applying PR on uniform-time noisy latents over-regularizes the latent space: the top PCA component degenerates into a near-constant channel that fails to capture spatial variations (i.e., over-smoothed latents). The linear head, optimized with additional weight decay, collapses to a trivial solution that arbitrarily inflates the latent norm to enlarge the logits.}
\label{tab:pr_classifier_ablation}

\begingroup
\small
\setlength{\tabcolsep}{6pt}
\renewcommand{\arraystretch}{1.18}
\definecolor{baselinegray}{gray}{0.94}

\resizebox{\columnwidth}{!}{%
\begin{tabular}{lccccc}
\toprule[1.2pt]
\multirow[c]{3}{*}{\textbf{Method}}
& \multicolumn{3}{c}{\textbf{Stage 1}}
& \multicolumn{2}{c}{\textbf{Stage 2}} \\
\cmidrule(lr){2-4}
\cmidrule(lr){5-6}
& \multirow[c]{2}{*}{rFID\,${\downarrow}$}
& \multirow[c]{2}{*}{PSNR\,${\uparrow}$}
& \multirow[c]{2}{*}{LPIPS\,${\downarrow}$}
& gFID\,${\downarrow}$
& gFID\,${\downarrow}$ \\
& & & & w/ CFG & w/o CFG \\
\midrule[0.6pt]

\rowcolor{baselinegray}
LDM
& 0.285 & \textbf{27.13} & \textbf{0.100}
& 4.56 & 18.79 \\

\midrule
LDM + PR (Cosine Head, Clean-only Latent)
& \textbf{0.279} & 26.75 & 0.103
& \textbf{3.24} & \textbf{17.11} \\

LDM + PR (Cosine Head, Uniform-time Noisy Latent)
& 0.457 & 26.83 & 0.110
& 14.02 & 41.63 \\

LDM + PR (Linear Head, Clean-only Latent)
& 0.319 & 26.64 & 0.108
& 21.55 & 64.18 \\

\bottomrule[1.2pt]
\end{tabular}%
}

\endgroup
\end{table}

\subsection{Ablation on Prototype Resolution}
\label{app:pr_resolution_ablation}

\begin{table}[h!]
\centering
\caption{\textbf{Ablation on prototype shape (reduced compute budget).} All variants use the default PR setting (cosine prototype classifier on clean-only latents) and differ only in the output shape of spatial average pooling applied before the classifier, where $16{\times}16$ corresponds to no pooling. Overall, prototypes with lower resolutions perform best: $1{\times}1$ prototypes (i.e., global average pooling) achieve the best rFID and gFID without CFG, and $2{\times}2$ prototypes achieve the best gFID with CFG, while further increasing prototype resolution consistently degrades generation quality. This suggests that reducing the spatial resolution of prototypes benefits generation, likely because it avoids imposing class-level layout constraints on the latent space.}
\label{tab:pr_resolution_ablation}

\begingroup
\small
\setlength{\tabcolsep}{6pt}
\renewcommand{\arraystretch}{1.18}
\definecolor{baselinegray}{gray}{0.94}

\resizebox{\columnwidth}{!}{%
\begin{tabular}{lccccc}
\toprule[1.2pt]
\multirow[c]{3}{*}{\textbf{Method}}
& \multicolumn{3}{c}{\textbf{Stage 1}}
& \multicolumn{2}{c}{\textbf{Stage 2}} \\
\cmidrule(lr){2-4}
\cmidrule(lr){5-6}
& \multirow[c]{2}{*}{rFID\,${\downarrow}$}
& \multirow[c]{2}{*}{PSNR\,${\uparrow}$}
& \multirow[c]{2}{*}{LPIPS\,${\downarrow}$}
& gFID\,${\downarrow}$
& gFID\,${\downarrow}$ \\
& & & & w/ CFG & w/o CFG \\
\midrule[0.6pt]

\rowcolor{baselinegray}
LDM
& 0.285 & \textbf{27.13} & \textbf{0.100}
& 4.56 & 18.79 \\

\midrule
LDM + PR ($1\times1$ Prototypes)
& \textbf{0.279} & 26.75 & 0.103
& 3.24 & \textbf{17.11} \\

LDM + PR ($2\times2$ Prototypes)
& 0.312 & 26.41 & 0.109
& \textbf{3.19} & 20.66 \\

LDM + PR ($4\times4$ Prototypes)
& 0.307 & 26.40 & 0.109
& 4.04 & 22.53 \\

LDM + PR ($16\times16$ Prototypes)
& 0.305 & 26.34 & 0.109
& 15.01 & 38.74 \\

\bottomrule[1.2pt]
\end{tabular}%
}
\endgroup
\end{table}

\newpage
\subsection{Ablation on Prototype Temperature}
\label{app:pr_temperature_ablation}

\begin{table}[h!]
\centering
\caption{\textbf{Ablation on the temperature of PR (reduced compute budget).}
$\tau$ scales the cosine-similarity logits of the prototype classifier, where a smaller $\tau$ yields a sharper softmax.
The default $\tau=0.1$ was fixed prior to this ablation and is used throughout the main text.
All PR variants improve over LDM both with and without CFG, and the differences among $\tau$ values are small compared with the gain from adding PR, indicating that $\tau=0.1$ is a reasonable default.
Adjusting $\tau$ (e.g., $\tau=0.3$ here) may bring further improvements, though the best value is likely setting-dependent.}
\label{tab:pr_temperature_ablation}

\begingroup
\small
\setlength{\tabcolsep}{6pt}
\renewcommand{\arraystretch}{1.18}
\definecolor{baselinegray}{gray}{0.94}

\resizebox{\columnwidth}{!}{%
\begin{tabular}{lccccc}
\toprule[1.2pt]
\multirow[c]{3}{*}{\textbf{Method}}
& \multicolumn{3}{c}{\textbf{Stage 1}}
& \multicolumn{2}{c}{\textbf{Stage 2}} \\
\cmidrule(lr){2-4}
\cmidrule(lr){5-6}
& \multirow[c]{2}{*}{rFID\,${\downarrow}$}
& \multirow[c]{2}{*}{PSNR\,${\uparrow}$}
& \multirow[c]{2}{*}{LPIPS\,${\downarrow}$}
& gFID\,${\downarrow}$
& gFID\,${\downarrow}$ \\
& & & & w/ CFG & w/o CFG \\
\midrule[0.6pt]

\rowcolor{baselinegray}
LDM
& 0.285 & \textbf{27.13} & \textbf{0.100}
& 4.56 & 18.79 \\

\midrule
LDM + PR ($\tau=0.03$)
& 0.299 & 26.76 & 0.104
& 3.27 & 17.61 \\

LDM + PR ($\tau=0.1$)
& \textbf{0.279} & 26.75 & 0.103
& 3.24 & 17.11 \\

LDM + PR ($\tau=0.3$)
& 0.291 & 26.76 & 0.103
& \textbf{3.14} & \textbf{16.24} \\

\bottomrule[1.2pt]
\end{tabular}%
}
\endgroup
\end{table}

\subsection{Ablation on Pseudo-label Granularity}
\label{app:psudo_granularity}

\begin{table}[h!]
\centering
\caption{\textbf{Ablation on DINO pseudo-label granularity (reduced compute budget).} This group of experiments uses a separate hyper-parameter setup, so the results are comparable within this table but not with other tables. Reconstruction is largely insensitive to the choice of targets. Unguided gFID degrades as $K$ increases, while guided gFID remains comparable across $K$ except for overly fine clusters. The nested variant (i.e., partitioning each of the 64 clusters into 16 sub-clusters) achieves the best generation among unsupervised targets (best unguided gFID and guided gFID close to GT labels), showing the feasibility of hierarchical PR.
}
\label{tab:dino_cluster_ablation}

\begingroup
\small
\setlength{\tabcolsep}{6pt}
\renewcommand{\arraystretch}{1.18}
\definecolor{baselinegray}{gray}{0.94}

\resizebox{\columnwidth}{!}{%
\begin{tabular}{lccccc}
\toprule[1.2pt]
\multirow[c]{3}{*}{\textbf{Labels / Clusters for PR}}
& \multicolumn{3}{c}{\textbf{Stage 1}}
& \multicolumn{2}{c}{\textbf{Stage 2}} \\
\cmidrule(lr){2-4}
\cmidrule(lr){5-6}
& \multirow[c]{2}{*}{rFID\,${\downarrow}$}
& \multirow[c]{2}{*}{PSNR\,${\uparrow}$}
& \multirow[c]{2}{*}{LPIPS\,${\downarrow}$}
& gFID\,${\downarrow}$
& gFID\,${\downarrow}$ \\
& & & & w/ CFG & w/o CFG \\
\midrule[0.6pt]

DINOv3-B $K=64$
& 0.264 & \textbf{28.09} & 0.096
& 3.01 & 14.07 \\

DINOv3-B $K=256$
& \textbf{0.254} & 28.00 & \textbf{0.095}
& 3.04 & 14.99 \\

DINOv3-B $K=1000$
& 0.294 & 27.59 & 0.101
& 2.99 & 17.60 \\

DINOv3-B $K=4096$
& 0.263 & 27.99 & 0.095
& 3.27 & 17.93 \\

DINOv3-B $K=64\times16$ (Nested)
& 0.264 & 27.99 & 0.096
& 2.94 & \textbf{13.63} \\

\midrule
\rowcolor{baselinegray}
GT Labels ($K=1000$)
& 0.263 & 27.82 & 0.098
& \textbf{2.87} & 14.60 \\

\bottomrule[1.2pt]
\end{tabular}%
}
\endgroup
\end{table}

\newpage
\subsection{Composability of PR and VA-VAE}
\label{app:compositionality}

\begin{table}[h!]
\centering
\caption{\textbf{Composability of PR and VA-VAE.} This group of experiments uses the same separate hyper-parameter setup as Tab.~\ref{tab:dino_cluster_ablation}, so the results are comparable with that table but not with the others. We apply PR on top of VA-VAE, which aligns to DINOv3-L features. Combining the two yields better generation both with and without CFG, at the cost of mildly degraded reconstruction, indicating that PR is complementary to external feature alignment for generation.}

\label{tab:pr_compositionality}

\begingroup
\small
\setlength{\tabcolsep}{6pt}
\renewcommand{\arraystretch}{1.18}
\definecolor{baselinegray}{gray}{0.94}

\resizebox{\columnwidth}{!}{%
\begin{tabular}{lccccc}
\toprule[1.2pt]
\multirow[c]{3}{*}{\textbf{Method}}
& \multicolumn{3}{c}{\textbf{Stage 1}}
& \multicolumn{2}{c}{\textbf{Stage 2}} \\
\cmidrule(lr){2-4}
\cmidrule(lr){5-6}
& \multirow[c]{2}{*}{rFID\,${\downarrow}$}
& \multirow[c]{2}{*}{PSNR\,${\uparrow}$}
& \multirow[c]{2}{*}{LPIPS\,${\downarrow}$}
& gFID\,${\downarrow}$
& gFID\,${\downarrow}$ \\
& & & & w/ CFG & w/o CFG \\
\midrule[0.6pt]

\rowcolor{baselinegray}
LDM
& 0.248 & 28.21 & 0.095
& 4.25 & 18.85 \\

\midrule
LDM + PR
& 0.263 & 27.82 & 0.098
& 2.87 & 14.60 \\

VAVAE (DINOv3-L)
& \textbf{0.229} & \textbf{28.27} & \textbf{0.092}
& 3.04 & 12.71 \\

VAVAE (DINOv3-L) + PR
& 0.282 & 27.55 & 0.101
& \textbf{2.57} & \textbf{10.89} \\

\bottomrule[1.2pt]
\end{tabular}%
}
\endgroup
\end{table}

\newpage
\section{Derivation of Equations and Claims in Main Text}
\label{app:cfg_derivations}

\subsection{Notation and setup}
\label{app:setup}

\paragraph{\textbf{Setup.}}
Let $C\in\{1,\ldots,K\}$ denote the class label and $Z\in\mathbb{R}^{d}$ the clean latent, with $\mathbb{E}\|Z\|^2<\infty$. Let $\epsilon\sim\mathcal{N}(0,I)$ be independent of $(Z,C)$. For a fixed $t\in(0,1)$, define $Z_t=tZ+(1-t)\epsilon$ and $v=Z-\epsilon$. We use $z$, $c$, and $z_t$ for realizations, write $q_j(z_t,t)=\Pr(C=j\mid Z_t=z_t,t)$, and abbreviate it as $q_j$ when the arguments are clear. The Bayes-optimal conditional and unconditional velocity predictors are denoted by $v_c^\star$ and $v_u^\star$, respectively, and $g_c^\star=v_c^\star-v_u^\star$. All expectations and entropies involving $Z_t$ are evaluated at the stated fixed time unless time is explicitly averaged over. Under the Gaussian class model, $Z\mid C=j\sim\mathcal{N}(\mu_j,\sigma^2I)$ with $\sigma^2>0$, and we write $s_t^2=t^2\sigma^2+(1-t)^2$. Class priors are fixed and positive, but need not be equal unless stated otherwise. For PR, the latent is viewed as a tensor $z\in\mathbb{R}^{d_c\times h\times w}$, with $d=d_chw$, pooled feature $\bar z=\mathrm{AvgPool}(z)$, and normalized features $\tilde z=\bar z/\|\bar z\|$ and $\tilde p_j=p_j/\|p_j\|$. These normalized vectors are assumed to have nonzero denominators. We reserve $q_j$ for the true posterior and $\pi_j$ for the prototype classifier. All logarithms are natural.

\subsection{Posterior-mean expression for CFG (Eq.~\ref{eq:cfg_posterior_mean})}
\label{app:cfg_posterior_mean}

\paragraph{\textbf{Derivation.}}
The interpolant gives $\epsilon=(z_t-tz)/(1-t)$, so the velocity target can be written as
\[
v=z-\epsilon=\frac{z-z_t}{1-t}.
\]
Under squared-error prediction, the Bayes-optimal predictors are conditional expectations. Since $z_t$ and $t$ are fixed within these expectations,
\[
v_c^\star=\frac{\mathbb{E}[Z\mid Z_t=z_t,t,C=c]-z_t}{1-t},
\qquad
v_u^\star=\frac{\mathbb{E}[Z\mid Z_t=z_t,t]-z_t}{1-t}.
\]
Subtracting cancels the common observed latent and yields
\begin{equation}
g_c^\star(z_t,t)=\frac{\mathbb{E}[Z\mid Z_t=z_t,t,C=c]-\mathbb{E}[Z\mid Z_t=z_t,t]}{1-t}.
\label{app:cfg_mean_identity}
\end{equation}
This identity does not require the Gaussian class model.

\subsection{Gaussian class geometry of CFG (Eq.~\ref{eq:cfg_class_geometry})}
\label{app:cfg_class_geometry}

\paragraph{\textbf{Derivation.}}
Under the Gaussian class model, the independent within-class variation and corruption noise give
\begin{equation}
Z_t\mid C=j\sim\mathcal{N}(t\mu_j,s_t^2I),
\qquad
s_t^2=t^2\sigma^2+(1-t)^2.
\label{app:gaussian_noisy_likelihood}
\end{equation}
Bayes' rule therefore gives
\[
q_j(z_t,t)=\frac{\Pr(C=j)\exp\!\left(-\frac{\|z_t-t\mu_j\|^2}{2s_t^2}\right)}{\sum_{k=1}^{K}\Pr(C=k)\exp\!\left(-\frac{\|z_t-t\mu_k\|^2}{2s_t^2}\right)}.
\]
Within class $j$, the clean posterior mean combines the Gaussian prior and corruption likelihood:
\[
p(z\mid z_t,t,C=j)\propto\exp\!\left(-\frac{\|z-\mu_j\|^2}{2\sigma^2}-\frac{\|z_t-tz\|^2}{2(1-t)^2}\right).
\]
Completing the square gives
\[
\begin{aligned}
\mathbb{E}[Z\mid Z_t=z_t,t,C=j]
&=\left(\frac{1}{\sigma^2}+\frac{t^2}{(1-t)^2}\right)^{-1}\left(\frac{\mu_j}{\sigma^2}+\frac{tz_t}{(1-t)^2}\right)\\
&=\frac{t\sigma^2}{s_t^2}z_t+\frac{(1-t)^2}{s_t^2}\mu_j.
\end{aligned}
\]
Averaging over the posterior class probabilities gives
\[
\mathbb{E}[Z\mid Z_t=z_t,t]=\frac{t\sigma^2}{s_t^2}z_t+\frac{(1-t)^2}{s_t^2}\sum_{j=1}^{K}q_j\mu_j.
\]
Substituting both posterior means into Eq.~\ref{app:cfg_mean_identity} yields
\[
\begin{aligned}
g_c^\star(z_t,t)
&=\frac{1-t}{s_t^2}\left(\mu_c-\sum_{j=1}^{K}q_j\mu_j\right)\\
&=\frac{1-t}{(1-t)^2+t^2\sigma^2}\sum_{j\neq c}q_j(\mu_c-\mu_j),
\end{aligned}
\]
where the last equality uses $\sum_jq_j=1$.

\subsection{Class separation under corruption (Eq.~\ref{eq:noisy_class_separation})}
\label{app:noisy_class_separation}

\paragraph{\textbf{Derivation.}}
By Eq.~\ref{app:gaussian_noisy_likelihood}, the corrupted class means are $t\mu_c$ and $t\mu_j$, with shared covariance $s_t^2I$. Their squared standardized distance is therefore
\[
\begin{aligned}
\delta_{cj}^2(t)
&=(t\mu_c-t\mu_j)^\top(s_t^2I)^{-1}(t\mu_c-t\mu_j)\\
&=\frac{t^2\|\mu_c-\mu_j\|^2}{t^2\sigma^2+(1-t)^2}.
\end{aligned}
\]

\subsection{Fixed-variance construction (Eq.~\ref{eq:variance_allocation})}
\label{app:variance_allocation}

\paragraph{\textbf{Verification.}}
Let $\mu_c=\sqrt{\alpha}\,\mu_c^{\mathrm{ref}}$ and $\sigma^2=1-\alpha$, where $\mathbb{E}_C[\mu_C^{\mathrm{ref}}]=0$ and $\mathbb{E}_C\|\mu_C^{\mathrm{ref}}\|^2=d$. Then $\mathbb{E}[Z]=0$, and the inter-class and intra-class variance contributions are, respectively,
\[
\mathbb{E}_C\|\mathbb{E}[Z\mid C]-\mathbb{E}[Z]\|^2=\alpha\,\mathbb{E}_C\|\mu_C^{\mathrm{ref}}\|^2=\alpha d,
\]
\[
\mathbb{E}_C\operatorname{tr}\operatorname{Cov}(Z\mid C)=d\sigma^2=(1-\alpha)d.
\]
The law of total variance therefore gives
\[
\operatorname{tr}\operatorname{Cov}(Z)=\alpha d+(1-\alpha)d=d.
\]
This reallocates a fixed total variance between class means and within-class variation.

\subsection{Posterior cross-entropy and conditional entropy (Eq.~\ref{eq:posterior_objective})}
\label{app:posterior_entropy}

\paragraph{\textbf{Derivation.}}
At fixed $t$, sampling $(Z,C,\epsilon)$ induces the joint distribution of $(Z_t,C)$. Conditioning first on $Z_t$ and using the true class posterior gives
\[
\begin{aligned}
\mathcal{L}_{\mathrm{post}}(t)
&=\mathbb{E}_{Z,C,\epsilon}\!\left[-\log q_C(tZ+(1-t)\epsilon,t)\right]\\
&=\mathbb{E}_{Z_t}\!\left[\mathbb{E}_{C\mid Z_t,t}\!\left[-\log q_C(Z_t,t)\right]\right]\\
&=\mathbb{E}_{Z_t}\!\left[-\sum_{j=1}^{K}q_j(Z_t,t)\log q_j(Z_t,t)\right]\\
&=H(C\mid Z_t).
\end{aligned}
\]
The equality relies on using the true posterior, rather than an arbitrary classifier.

\subsection{Posterior gradient and CFG (Eq.~\ref{eq:posterior_objective_gradient})}
\label{app:posterior_objective_gradient}

\paragraph{\textbf{Derivation.}}
We apply Tweedie's formula under Gaussian forward corruption without assuming a Gaussian class model for the clean latent distribution.

The corruption likelihood is $p(z_t\mid z,t)=\mathcal{N}(z_t;tz,(1-t)^2I)$. Differentiating its marginal with respect to $z_t$ gives
\[
\begin{aligned}
\nabla_{z_t}\log p(z_t\mid C=c,t)
&=\frac{\mathbb{E}_{Z\mid C=c}[\nabla_{z_t}p(z_t\mid Z,t)]}{p(z_t\mid C=c,t)}\\
&=\mathbb{E}\!\left[\left.\frac{tZ-z_t}{(1-t)^2}\right|Z_t=z_t,t,C=c\right]\\
&=\frac{t\,\mathbb{E}[Z\mid Z_t=z_t,t,C=c]-z_t}{(1-t)^2}.
\end{aligned}
\]
The same calculation without conditioning on the class yields
\[
\nabla_{z_t}\log p(z_t\mid t)=\frac{t\,\mathbb{E}[Z\mid Z_t=z_t,t]-z_t}{(1-t)^2}.
\]
By Bayes' rule, the class prior contributes no derivative with respect to $z_t$. Subtracting the two scores and using Eq.~\ref{app:cfg_mean_identity}, we obtain
\[
\begin{aligned}
\nabla_{z_t}\log q_c(z_t,t)
&=\nabla_{z_t}\log p(z_t\mid C=c,t)-\nabla_{z_t}\log p(z_t\mid t)\\
&=\frac{t}{(1-t)^2}\left(\mathbb{E}[Z\mid Z_t=z_t,t,C=c]-\mathbb{E}[Z\mid Z_t=z_t,t]\right)\\
&=\frac{t}{1-t}g_c^\star(z_t,t).
\end{aligned}
\]
Finally, hold $q_c$, $t$, and $\epsilon$ fixed and differentiate through $z_t=tz+(1-t)\epsilon$:
\[
\nabla_z\log q_c(tz+(1-t)\epsilon,t)=t\nabla_{z_t}\log q_c(z_t,t)=\frac{t^2}{1-t}g_c^\star(z_t,t).
\]
This is the per-sample negative cross-entropy gradient with a fixed posterior function; it does not include changes in that function induced by changing the underlying latent distribution. The identity does not require the Gaussian class model.

\subsection{Monotonic posterior loss under variance allocation (Eqs.~\ref{eq:variance_allocation} and~\ref{eq:posterior_objective})}
\label{app:posterior_objective}

\paragraph{\textbf{Derivation.}}
Under the Gaussian family in Appendix~\ref{app:variance_allocation}, keep the class priors and reference means $\{\mu_c^{\mathrm{ref}}\}$ fixed, and write
\[
Z=\sqrt{\alpha}\,\mu_C^{\mathrm{ref}}+\sqrt{1-\alpha}\,\eta,
\qquad \alpha\in(0,1),
\]
where $\eta$ and the corruption noise $\epsilon$ are independent standard Gaussian vectors, both independent of $C$.
For any fixed $t\in(0,1]$, the corrupted latent is
\[
Z_t=t\sqrt{\alpha}\,\mu_C^{\mathrm{ref}}
+t\sqrt{1-\alpha}\,\eta+(1-t)\epsilon.
\]
Since $t^2(1-\alpha)+(1-t)^2>0$ throughout this domain, combining the two Gaussian noise terms and rescaling the observation gives the equivalent channel
\[
Y_\alpha
:=\frac{Z_t}{\sqrt{t^2(1-\alpha)+(1-t)^2}}
=\sqrt{\gamma_t(\alpha)}\,\mu_C^{\mathrm{ref}}+\xi,
\qquad
\gamma_t(\alpha)
:=\frac{t^2\alpha}{t^2(1-\alpha)+(1-t)^2},
\]
where $\xi\sim\mathcal{N}(0,I)$ is independent of $C$.
For each $\alpha$, the rescaling is invertible, so
$H(C\mid Z_t)=H(C\mid Y_\alpha)$.
Moreover,
\[
\frac{d\gamma_t(\alpha)}{d\alpha}
=
\frac{t^2\bigl(t^2+(1-t)^2\bigr)}
{\bigl(t^2(1-\alpha)+(1-t)^2\bigr)^2}
>0.
\]

To compare $0<\alpha_1<\alpha_2<1$, write
$\gamma_i=\gamma_t(\alpha_i)$, so that $0<\gamma_1<\gamma_2$, and construct
\[
Y_2=\sqrt{\gamma_2}\,\mu_C^{\mathrm{ref}}+\xi_2,
\qquad
\widehat Y_1
=\sqrt{\frac{\gamma_1}{\gamma_2}}\,Y_2
+\sqrt{1-\frac{\gamma_1}{\gamma_2}}\,\xi',
\]
where $\xi_2$ and $\xi'$ are independent standard Gaussian vectors, both independent of $C$.
Conditional on $C$, $\widehat Y_1$ is Gaussian with mean
$\sqrt{\gamma_1}\,\mu_C^{\mathrm{ref}}$ and covariance $I$.
Thus $(C,\widehat Y_1)$ has the same distribution as $(C,Y_{\alpha_1})$.
Moreover, $C\to Y_2\to\widehat Y_1$ is a Markov chain.
The data-processing inequality thus gives
\[
I(C;Y_{\alpha_1})
=I(C;\widehat Y_1)
\leq I(C;Y_2)
=I(C;Y_{\alpha_2}).
\]
Since the class priors and hence $H(C)$ are fixed,
\[
H(C\mid Y_{\alpha_2})
=H(C)-I(C;Y_{\alpha_2})
\leq H(C)-I(C;Y_{\alpha_1})
=H(C\mid Y_{\alpha_1}).
\]
Using the entropy identity in Appendix~\ref{app:posterior_entropy}, we conclude that $\mathcal{L}_{\mathrm{post}}(t)$ is non-increasing in $\alpha$ at every fixed $t\in(0,1]$.

In particular, at $t=1$ we have $Z_t=Z$ and
\[
Y_\alpha=\frac{Z}{\sqrt{1-\alpha}},
\qquad
\gamma_1(\alpha)=\frac{\alpha}{1-\alpha}.
\]
The rescaling remains invertible for $\alpha\in(0,1)$, so the same argument directly establishes that the clean posterior loss $H(C\mid Z)$ is non-increasing in $\alpha$.

\subsection{Prototype posterior and class-contrast gradient (Eqs.~\ref{eq:pr} and~\ref{eq:pr_gradient})}
\label{app:pr}

\paragraph{\textbf{Gaussian posterior form.}}
For a clean latent under the Gaussian class model, Bayes' rule and expansion of the squared distance give
\[
q_j(z,1)=\frac{\exp\!\left(\frac{\mu_j^\top z}{\sigma^2}-\frac{\|\mu_j\|^2}{2\sigma^2}+\log\Pr(C=j)\right)}{\sum_{k=1}^{K}\exp\!\left(\frac{\mu_k^\top z}{\sigma^2}-\frac{\|\mu_k\|^2}{2\sigma^2}+\log\Pr(C=k)\right)}.
\]
With equal priors and equal-norm class means, the bias terms are common across classes and cancel:
\[
q_j(z,1)=\frac{\exp(\mu_j^\top z/\sigma^2)}{\sum_{k=1}^{K}\exp(\mu_k^\top z/\sigma^2)}.
\]
PR adopts this softmax inner-product form on normalized pooled features, with learnable normalized prototypes and temperature $\tau>0$. Its posterior and cross-entropy are defined as
\[
\pi_j(z)=\frac{\exp(\tilde z^\top\tilde p_j/\tau)}{\sum_{k=1}^{K}\exp(\tilde z^\top\tilde p_k/\tau)},
\qquad
\mathcal{L}_{\mathrm{PR}}=-\log\pi_c(z).
\]
This surrogate matches the Gaussian discriminant form without assuming that normalized pooled features themselves follow the original Gaussian class model.

\paragraph{\textbf{Gradient derivation.}}
Expand the cross-entropy as
\[
\mathcal{L}_{\mathrm{PR}}=-\frac{\tilde z^\top\tilde p_c}{\tau}+\log\sum_{k=1}^{K}\exp(\tilde z^\top\tilde p_k/\tau).
\]
Holding the prototypes fixed and differentiating with respect to the normalized feature gives
\[
\begin{aligned}
\nabla_{\tilde z}\mathcal{L}_{\mathrm{PR}}
&=-\frac{\tilde p_c}{\tau}+\frac{\sum_{j=1}^{K}\exp(\tilde z^\top\tilde p_j/\tau)\,\tilde p_j/\tau}{\sum_{k=1}^{K}\exp(\tilde z^\top\tilde p_k/\tau)}\\
&=\frac{1}{\tau}\left(\sum_{j=1}^{K}\pi_j\tilde p_j-\tilde p_c\right).
\end{aligned}
\]
Using $\sum_j\pi_j=1$, we obtain
\[
\begin{aligned}
-\nabla_{\tilde z}\mathcal{L}_{\mathrm{PR}}
&=\frac{1}{\tau}\left(\tilde p_c-\sum_{j=1}^{K}\pi_j\tilde p_j\right)\\
&=\frac{1}{\tau}\sum_{j\neq c}\pi_j(\tilde p_c-\tilde p_j).
\end{aligned}
\]
This has the same posterior-weighted class-contrast structure as the expression in Appendix~\ref{app:cfg_class_geometry}, with normalized prototypes and classifier probabilities replacing class means and true posterior probabilities.
For completeness, we derive the gradient with respect to the clean
latent. With $\bar z=N^{-1}\sum_{s=1}^{N}z_s$ and $N=hw$,
backpropagation through normalization and pooling gives
\[
-\nabla_{z_s}\mathcal{L}_{\mathrm{PR}}
=
\frac{I-\tilde z\tilde z^\top}{N\tau\|\bar z\|}
\sum_{j\neq c}\pi_j(\tilde p_c-\tilde p_j),
\qquad s=1,\ldots,N.
\]
Thus, the class contrast is projected onto the tangent space at $\tilde z$, rescaled, and distributed equally across spatial locations.
The resulting supervision promotes class discrimination in the pooled representation while leaving spatial residuals $z_s-\bar z$ unconstrained.

\newpage
\section{Complementary Perspectives on Classifier-Free Guidance}
\label{app:cfg_perspectives}

We examine CFG from three complementary perspectives: the irreducible uncertainty of velocity prediction, the error in a learned guidance correction, and the estimation accuracy attainable under a controlled model. The first isolates the role of latent geometry, the second separates conditional error contrasts from branch consistency, and the third connects these quantities through an explicit estimator. Throughout, $Z_t=tZ+(1-t)\epsilon$, $v=Z-\epsilon$, and $t\in(0,1)$ is fixed, with $\epsilon\sim\mathcal{N}(0,I)$ independent of $(Z,C)$. Expectations are taken over the forward process generated by $(Z,C,\epsilon)$. We use $q_j=\Pr(C=j\mid Z_t,t)$, $v_j^\star=\mathbb{E}[v\mid Z_t,t,C=j]$, $v_u^\star=\mathbb{E}[v\mid Z_t,t]$, and $g_c^\star=v_c^\star-v_u^\star$.

\subsection{Class Geometry and Irreducible Velocity Uncertainty}
\label{app:cfg_velocity_uncertainty}

\paragraph{\textbf{Takeaway and setup.}}
Under the Gaussian class model, reallocating a fixed total variance toward class means strictly reduces the conditional velocity Bayes risk and does not increase the unconditional risk. Assume $Z\mid C=c\sim\mathcal{N}(\mu_c,\sigma^2I)$ and write $s_t^2=(1-t)^2+t^2\sigma^2$. For the geometric comparison, fix the class priors and reference means satisfying $\mathbb{E}_C[\mu_C^{\mathrm{ref}}]=0$ and $\mathbb{E}_C\|\mu_C^{\mathrm{ref}}\|^2=d$, and set $\mu_c=\sqrt{\alpha}\,\mu_c^{\mathrm{ref}}$ and $\sigma^2=1-\alpha$, with $\alpha\in(0,1)$.

\paragraph{\textbf{Velocity Bayes risks.}}
The conditional and unconditional irreducible errors satisfy
\begin{equation}
\begin{aligned}
\mathcal{R}_{\mathrm{cond}}(t)&:=\mathbb{E}\|v-v_C^\star\|^2=\frac{d\sigma^2}{s_t^2},\\
\mathcal{R}_{\mathrm{uncond}}(t)&:=\mathbb{E}\|v-v_u^\star\|^2=\mathcal{R}_{\mathrm{cond}}(t)+\mathbb{E}\|g_C^\star\|^2.
\end{aligned}
\label{app:cfg_velocity_bayes_risks}
\end{equation}
To obtain the first identity, Gaussian conditioning gives
\[
\operatorname{Cov}(Z\mid Z_t,t,C)=\left(\frac{1}{\sigma^2}+\frac{t^2}{(1-t)^2}\right)^{-1}I=\frac{\sigma^2(1-t)^2}{s_t^2}I.
\]
Since $v=(Z-Z_t)/(1-t)$, its conditional covariance is $\sigma^2I/s_t^2$, whose trace gives $\mathcal{R}_{\mathrm{cond}}$. For the second identity, write $v-v_u^\star=(v-v_C^\star)+g_C^\star$. The cross term vanishes because $\mathbb{E}[v-v_C^\star\mid Z_t,t,C]=0$.

\paragraph{\textbf{Effect of variance allocation.}}
For any $0<\alpha_1<\alpha_2<1$, we have
\begin{equation}
\begin{aligned}
\mathcal{R}_{\mathrm{cond}}(t;\alpha_2)&<\mathcal{R}_{\mathrm{cond}}(t;\alpha_1),\\
\mathcal{R}_{\mathrm{uncond}}(t;\alpha_2)&\leq\mathcal{R}_{\mathrm{uncond}}(t;\alpha_1).
\end{aligned}
\label{app:cfg_variance_reallocation}
\end{equation}
The conditional statement follows directly from
\[
\frac{\partial\mathcal{R}_{\mathrm{cond}}(t;\alpha)}{\partial\alpha}=-\frac{d(1-t)^2}{\bigl((1-t)^2+t^2(1-\alpha)\bigr)^2}<0.
\]
For the unconditional statement, couple the two models as follows. Let $Z_2$ follow the family at $\alpha_2$, let $\epsilon_2\sim\mathcal{N}(0,I)$ be independent, and take independent standard Gaussians $\eta_0,\epsilon_0$ independent of $(Z_2,C,\epsilon_2)$. Set
\[
\rho=\sqrt{\frac{\alpha_1}{\alpha_2}},\qquad Z_1=\rho Z_2+\sqrt{1-\rho^2}\,\eta_0,\qquad \epsilon_1=\rho\epsilon_2+\sqrt{1-\rho^2}\,\epsilon_0.
\]
Then $Z_1\mid C=c\sim\mathcal{N}(\sqrt{\alpha_1}\mu_c^{\mathrm{ref}},(1-\alpha_1)I)$, and $\epsilon_1$ is independent of $(Z_1,C)$. Write $Y_i=tZ_i+(1-t)\epsilon_i$ and $V_i=Z_i-\epsilon_i$ for $i=1,2$. Remove a common linear predictor by defining
\[
A_t=t^2+(1-t)^2,\qquad \kappa_t=\frac{2t-1}{A_t},\qquad r_i=V_i-\kappa_tY_i.
\]
The fixed total variance implies
\[
\mathbb{E}\|Y_i\|^2=dA_t,\qquad \mathbb{E}[V_i^\top Y_i]=d(2t-1),\qquad \mathbb{E}\|V_i\|^2=2d,
\]
and hence $\mathbb{E}\|r_i\|^2=d/A_t$ for both models. Moreover,
\[
Y_1=\rho Y_2+\sqrt{1-\rho^2}\,Y_0,\qquad r_1=\rho r_2+\sqrt{1-\rho^2}\,r_0,
\]
where $Y_0=t\eta_0+(1-t)\epsilon_0$ and $r_0=\eta_0-\epsilon_0-\kappa_tY_0$. These two auxiliary variables are jointly Gaussian with zero cross-covariance, so $r_0$ is independent of $Y_0$, and both are independent of $(Y_2,r_2)$. Therefore,
\[
\mathbb{E}[r_1\mid Y_1]=\rho\,\mathbb{E}\!\left[\mathbb{E}[r_2\mid Y_2]\mid Y_1\right],\qquad \mathbb{E}\|\mathbb{E}[r_1\mid Y_1]\|^2\leq\rho^2\mathbb{E}\|\mathbb{E}[r_2\mid Y_2]\|^2.
\]
Finally, subtracting a known function of the observation does not change the Bayes risk. Orthogonality of conditional expectation gives
\[
\begin{aligned}
\mathcal{R}_{\mathrm{uncond}}(t;\alpha_1)
&=\frac{d}{A_t}-\mathbb{E}\|\mathbb{E}[r_1\mid Y_1]\|^2\\
&\geq\frac{d}{A_t}-\rho^2\mathbb{E}\|\mathbb{E}[r_2\mid Y_2]\|^2\\
&\geq\frac{d}{A_t}-\mathbb{E}\|\mathbb{E}[r_2\mid Y_2]\|^2=\mathcal{R}_{\mathrm{uncond}}(t;\alpha_2).
\end{aligned}
\]

\paragraph{\textbf{Connection to diffusability.}}
Velocity irreducible variance has been studied as a statistical indicator of latent diffusability~\citep{zhong2026diffusing}. The conditional risk in Eq.~\ref{app:cfg_velocity_bayes_risks} is the isotropic Gaussian instance of that quantity. Equation~\ref{app:cfg_variance_reallocation} connects this perspective to class geometry: at a fixed latent scale, increasing the inter-class variance fraction reduces velocity-target uncertainty in the conditional branch and preserves or reduces it in the unconditional branch.

\newpage
\subsection{Conditional Error Contrasts and Branch Consistency}
\label{app:cfg_correction_error}

\paragraph{\textbf{Takeaway and setup.}}
The error in a learned CFG correction separates exactly into differences between conditional prediction errors and a consistency term relating the two branches. This decomposition holds without Gaussian assumptions. Let $e_j=\widehat v_j-v_j^\star$ and $e_u=\widehat v_u-v_u^\star$, and define $\widehat g_c=\widehat v_c-\widehat v_u$. Treat the learned predictors as fixed when averaging over the forward distribution.

\paragraph{\textbf{Exact correction-error decomposition.}}
The expected correction error is
\begin{equation}
\begin{aligned}
\mathcal{E}_g(t)&:=\mathbb{E}\|\widehat g_C-g_C^\star\|^2=\mathcal{E}_g^{(1)}(t)+\mathcal{E}_g^{(2)}(t),\\
\mathcal{E}_g^{(1)}(t)&:=\frac12\mathbb{E}\sum_{i,j}q_iq_j\|e_i-e_j\|^2,\\
\mathcal{E}_g^{(2)}(t)&:=\mathbb{E}\left\|\widehat v_u-\sum_jq_j\widehat v_j\right\|^2.
\end{aligned}
\label{app:cfg_error_decomposition}
\end{equation}
To see this, the tower property gives $v_u^\star=\sum_jq_jv_j^\star$. Define $\bar e=\sum_jq_je_j$ and $\Delta_u=\widehat v_u-\sum_jq_j\widehat v_j=e_u-\bar e$. Then
\[
\widehat g_C-g_C^\star=e_C-e_u=(e_C-\bar e)-\Delta_u.
\]
Conditional on $Z_t$, we have $\mathbb{E}[e_C-\bar e\mid Z_t,t]=0$, so the cross term vanishes. The remaining conditional variance satisfies
\[
\sum_jq_j\|e_j-\bar e\|^2=\sum_jq_j\|e_j\|^2-\|\bar e\|^2=\frac12\sum_{i,j}q_iq_j\|e_i-e_j\|^2,
\]
which proves Eq.~\ref{app:cfg_error_decomposition}. The first term measures error differences across jointly plausible classes; any component shared by all $e_j$ cancels. The second measures agreement between the learned unconditional prediction and the posterior-weighted conditional predictions.

\paragraph{\textbf{Posterior weighting and prediction uncertainty.}}
Define the expected posterior ambiguity as
\[
U_t:=\mathbb{E}\!\left[1-\sum_jq_j^2\right]=\mathbb{E}\sum_{i\neq j}q_iq_j.
\]
If $\|e_i-e_j\|\leq B$ uniformly, then $\mathcal{E}_g^{(1)}(t)\leq B^2U_t/2$. Thus, for a common error-difference bound, lower posterior ambiguity tightens the bound on the conditional contrast term. These learned errors are distinct from the irreducible uncertainty in Appendix~\ref{app:cfg_velocity_uncertainty}; for example, conditional prediction obeys
\[
\mathbb{E}\|v-\widehat v_C\|^2=\mathcal{R}_{\mathrm{cond}}(t)+\mathbb{E}\|e_C\|^2.
\]
Here the first term is determined by the data distribution, while the second measures approximation of its oracle predictor.

\newpage
\subsection{Geometry-Dependent Estimation Error under Consistent Branches}
\label{app:cfg_controlled_estimation}

\paragraph{\textbf{Takeaway and setup.}}
Under a controlled Gaussian estimator, within-class compactness and low posterior ambiguity jointly reduce correction and guided-field estimation errors. Use the same fixed-variance family as in Appendix~\ref{app:cfg_velocity_uncertainty}. Assume that the common linear term of the conditional predictors is known, estimate each class offset from $n$ independent residual samples, and construct the unconditional estimate using the true posterior weights. Estimation samples are independent across classes and independent of the test sample. Expectations below also average over these estimation samples, with $n$ and the guidance scale $\omega\geq1$ held fixed across the family.

\paragraph{\textbf{Estimating the class offsets.}}
The Gaussian conditional predictor has the form
\[
v_j^\star(z_t,t)=a_tz_t+h_j,\qquad a_t=\frac{t\sigma^2-(1-t)}{s_t^2},\qquad h_j=\frac{1-t}{s_t^2}\mu_j.
\]
For class $j$, a velocity residual $v-a_tZ_t$ has mean $h_j$ and covariance $\sigma^2I/s_t^2$: subtracting $h_j$ gives $v-v_j^\star$, whose conditional mean is zero and whose conditional covariance was computed in Appendix~\ref{app:cfg_velocity_uncertainty}. Estimate
\[
\widehat h_j=\frac1n\sum_{\ell=1}^{n}\bigl(v_{j\ell}-a_tZ_{t,j\ell}\bigr),\qquad \widehat v_j(z_t,t)=a_tz_t+\widehat h_j,\qquad \widehat v_u=\sum_jq_j\widehat v_j.
\]
Consequently, $e_j=\widehat h_j-h_j$ is independent of the test sample and satisfies
\[
\mathbb{E}[e_j]=0,\qquad \mathbb{E}\|e_j\|^2=\frac{d\sigma^2}{ns_t^2}=\frac{\mathcal{R}_{\mathrm{cond}}(t)}{n},\qquad \mathbb{E}[e_i^\top e_j]=0\quad(i\neq j).
\]

\paragraph{\textbf{Correction and guided-field errors.}}
The construction makes $\mathcal{E}_g^{(2)}(t)=0$ in Eq.~\ref{app:cfg_error_decomposition}. Since $\mathbb{E}\|e_i-e_j\|^2=2\mathcal{R}_{\mathrm{cond}}(t)/n$ for $i\neq j$, its first term gives the correction error below. For guidance, define $v_{\mathrm{cfg}}^\star=v_u^\star+\omega(v_C^\star-v_u^\star)$ and $\widehat v_{\mathrm{cfg}}=\widehat v_u+\omega(\widehat v_C-\widehat v_u)$. Then
\[
\widehat v_{\mathrm{cfg}}-v_{\mathrm{cfg}}^\star=\omega e_C+(1-\omega)\bar e=\sum_j\bigl(\omega\mathbf{1}_{\{j=C\}}+(1-\omega)q_j\bigr)e_j.
\]
Independence and zero-mean estimation errors remove cross terms. Averaging the squared coefficients over $C\mid Z_t$ gives
\[
\sum_cq_c\sum_j\bigl(\omega\mathbf{1}_{\{j=c\}}+(1-\omega)q_j\bigr)^2=\omega^2+(1-\omega^2)\sum_jq_j^2.
\]
Thus, the two estimation errors are
\begin{equation}
\begin{aligned}
\mathcal{E}_g(t)&=\frac{\mathcal{R}_{\mathrm{cond}}(t)}{n}\,U_t,\\
\mathcal{E}_\omega(t)&:=\mathbb{E}\|\widehat v_{\mathrm{cfg}}-v_{\mathrm{cfg}}^\star\|^2=\frac{\mathcal{R}_{\mathrm{cond}}(t)}{n}\bigl[1+(\omega^2-1)U_t\bigr].
\end{aligned}
\label{app:cfg_estimation_errors}
\end{equation}

\paragraph{\textbf{Monotonicity under variance allocation.}}
It remains to show that $U_t$ is non-increasing in $\alpha$. Rescaling the noisy latent gives an equivalent observation $Y_\alpha=\sqrt{\gamma_t(\alpha)}\mu_C^{\mathrm{ref}}+\xi$, where $\xi\sim\mathcal{N}(0,I)$ is independent of $C$ and
\[
\gamma_t(\alpha)=\frac{t^2\alpha}{t^2(1-\alpha)+(1-t)^2}.
\]
This signal-to-noise ratio increases with $\alpha$. For $\alpha_1<\alpha_2$, the Gaussian channel coupling in Appendix~\ref{app:posterior_objective} constructs observations with the corresponding marginals and $C\to Y_{\alpha_2}\to Y_{\alpha_1}$. Writing $\mathbf q_i=\Pr(C=\cdot\mid Y_{\alpha_i})$ for their posterior probability vectors, the tower property and Jensen's inequality give
\[
\mathbf q_1=\mathbb{E}[\mathbf q_2\mid Y_{\alpha_1}],\qquad \mathbb{E}\|\mathbf q_1\|^2\leq\mathbb{E}\|\mathbf q_2\|^2,
\]
and hence $U_t(\alpha_2)\leq U_t(\alpha_1)$. Together with the decrease of $\mathcal{R}_{\mathrm{cond}}$ in Eq.~\ref{app:cfg_variance_reallocation}, Eq.~\ref{app:cfg_estimation_errors} shows that both $\mathcal{E}_g(t)$ and $\mathcal{E}_\omega(t)$ are non-increasing in $\alpha$. Within this estimator, allocating more variance to class means lowers the cost of estimating each conditional predictor and reduces the posterior ambiguity that weights class contrasts.

\end{document}